\def\year{2021}\relax
\documentclass[letterpaper]{article} 
\usepackage{aaai21}  
\usepackage{times}  
\usepackage{helvet} 
\usepackage{courier}  
\usepackage[hyphens]{url}  
\usepackage{graphicx} 
\usepackage{graphicx}  
\usepackage{natbib}  
\usepackage{caption} 
\usepackage[switch]{lineno}

\usepackage{xcolor}
\usepackage{times}
\usepackage{paralist}
\usepackage{url}
\usepackage{comment}
\usepackage[flushleft]{threeparttable}
\usepackage{amsmath}
\usepackage{subfig}
\usepackage{tabularx}
\usepackage{mathrsfs}
\usepackage{algorithm}
\usepackage{enumitem}
\usepackage[multiple]{footmisc}
\usepackage{afterpage}

\usepackage{bm}
\usepackage{pbox}
\usepackage{graphicx}
\usepackage{multirow, makecell}
\usepackage{booktabs}
\usepackage{amssymb}
\DeclareMathOperator*{\softmax}{softmax}

\usepackage{todonotes}

\usepackage[utf8]{inputenc}

\newcommand{\bone}{\textsf{CL\_Trans\_GCN}}
\newcommand{\btwo}{\textsf{CL\_GCN}}
\newcommand{\bthree}{\textsf{CL\_RNN}}
\newcommand{\bfour}{\textsf{Transformer}}
\newcommand{\bfive}{\textsf{Transformer\_RPR}}

\definecolor{Blue}{RGB}{0, 0, 255}
\definecolor{Orange}{RGB}{244, 101, 66}
\definecolor{Purple}{RGB}{128, 0, 128}
\definecolor{Red}{RGB}{255, 0, 0}
\definecolor{forestgreen(web)}{rgb}{0.13, 0.55, 0.13}
\definecolor{Fuchsia}{HTML}{FF00FF}
\definecolor{green}{rgb}{0.0, 0.8, 0.0}

\usepackage{amsthm}

\title{GATE: Graph Attention Transformer Encoder for Cross-lingual\\Relation and Event Extraction}

\author{
    Wasi Uddin Ahmad, Nanyun Peng, Kai-Wei Chang \\
}
\affiliations{
    University of California, Los Angeles\\

    {\{wasiahmad, violetpeng, kwchang\}@cs.ucla.edu}\\

}

\begin{document}

\setlength{\abovedisplayskip}{4pt}
\setlength{\belowdisplayskip}{4pt}

\maketitle

\begin{abstract}
Recent progress in cross-lingual relation and event extraction use graph convolutional networks (GCNs) with universal dependency parses to learn language-agnostic sentence representations such that models trained on one language can be applied to other languages. However, GCNs struggle to model words with long-range dependencies or are not directly connected in the dependency tree. To address these challenges, we propose to utilize the self-attention mechanism where we explicitly fuse structural information to learn the dependencies between words with different syntactic distances. We introduce GATE, a {\bf G}raph {\bf A}ttention {\bf T}ransformer {\bf E}ncoder, and test its cross-lingual transferability on relation and event extraction tasks. We perform experiments on the ACE05 dataset that includes three typologically different languages: English, Chinese, and Arabic. The evaluation results show that GATE outperforms three recently proposed methods by a large margin. Our detailed analysis reveals that due to the reliance on syntactic dependencies, GATE produces robust representations that facilitate transfer across languages.
\end{abstract}

\section{Introduction}

\begin{figure}[t]
\centering
\vspace{-2mm}
\includegraphics[width=0.9\linewidth]{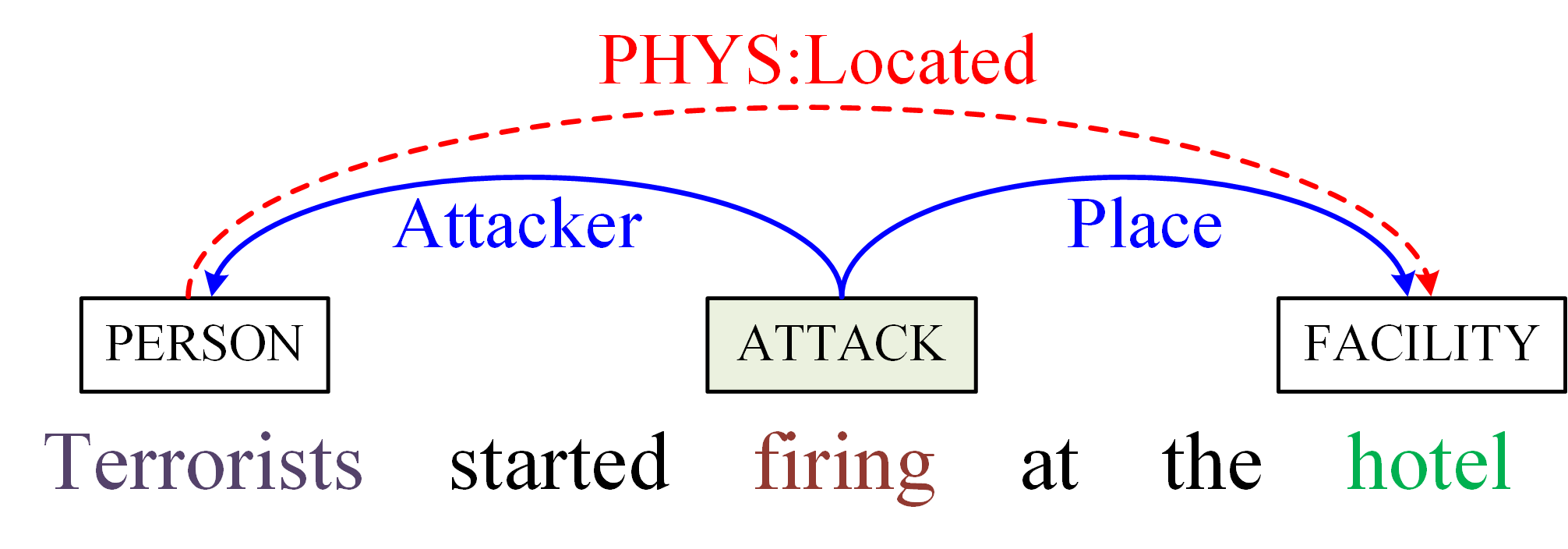}
\vspace{-2mm}
\caption{
A relation (red dashed) between two entities and an event of type \emph{Attack} (triggered by ``firing'') including two arguments  and their role labels (blue) are highlighted.
}
\label{fig:example}
\vspace{-2mm}
\end{figure}

Relation and event extraction are two challenging information extraction (IE) tasks; wherein a model learns to identify semantic relationships between entities and events in narratives. 
They provide useful information for many natural language processing (NLP) applications such as knowledge graph completion \cite{lin2015learning} and question answering \cite{chen-etal-2019-uhop}.
Figure~\ref{fig:example} gives an example of relation and event extraction tasks.
Recent advances in cross-lingual transfer learning approaches for relation and event extraction learns a universal encoder that produces language-agnostic contextualized representations so the model learned on one language can easily transfer to others. 
Recent works \cite{huang-etal-2018-zero, subburathinam2019cross} suggested embedding {\em universal dependency structure} into contextual representations improves cross-lingual transfer for IE.

There are a couple of advantages of leveraging dependency structures.
First, the syntactic distance between two words\footnote{The shortest path in the dependency graph structure.} in a sentence is typically smaller than the sequential distance. 
For example, in the sentence \emph{A fire in a Bangladeshi garment factory has left at least 37 people dead and 100 hospitalized}, the sequential and syntactic distance between ``fire'' and ``hospitalized'' is 15 and 4, respectively.
Therefore, encoding syntax structure helps capture long-range dependencies \cite{liu-etal-2018-jointly}.
Second, languages have different word order, e.g., adjectives precede or follow nouns as (``red apple'') in English or (``pomme rouge'') in French.
Thus, processing sentences sequentially suffers from the word order difference issue \cite{ahmad-etal-2019-difficulties}, while modeling dependency structures can mitigate the problem in cross-lingual transfer \cite{liu2019neural}.

A common way to leverage dependency structures for cross-lingual NLP tasks is using universal dependency parses.\footnote{https://universaldependencies.org/}
A large pool of recent works in IE \cite{liu-etal-2018-jointly, zhang-etal-2018-graph, subburathinam2019cross, fu2019graphrel, sun2019joint, liu2019neural} employed Graph Convolutional Networks (GCNs) \cite{kipf2017semi} to learn sentence representations based on their universal dependency parses,
where a $k$-layers GCN aggregates information of words that are $k$ hop away.
Such a way of embedding structure may hinder cross-lingual transfer when the source and target languages have different path length distributions among words (see Table \ref{table:dist_stat}).
Presumably, a two-layer GCN would work well on English but may not transfer well to Arabic.
Moreover, GCNs have shown to perform poorly in modeling long-distance dependencies or disconnected words in the dependency tree \cite{zhang-etal-2019-aspect, tang2020dependency}.
In contrast, the self-attention mechanism \cite{vaswani2017attention} is capable of capturing long-range dependencies.
Consequently, a few recent studies proposed dependency-aware self-attention and found effective for machine translation \cite{deguchi-etal-2019-dependency, bugliarello-okazaki-2020-enhancing}.
The key idea is to allow attention between connected words in the dependency tree and gradually aggregate information across layers.
However, IE tasks are relatively low-resource (the number of annotated documents available for training is small), and thus stacking more layers is not feasible. 
Besides, our preliminary analysis indicates that syntactic distance between entities could characterize certain relation and event types.\footnote{In ACE 2005 dataset, the relation type {\tt PHYS:Located} exists among {\tt \{PER, ORG, LOC, FAC, GPE\}} entities. The average syntactic distance in English and Arabic sentences among {\tt PER} and any of the {\tt \{LOC, FAC, GPE\}} entities are approx. 2.8 and 4.2, while the distance between  {\tt PER} and {\tt ORG} is 3.3 and 1.5.} 
Hence, we propose to allow attention between all words but use the {\em pairwise syntactic distances} to weigh the attention.



We introduce a {\bf G}raph {\bf A}ttention {\bf T}ransformer {\bf E}ncoder (GATE) that utilizes self-attention \cite{vaswani2017attention} to learn structured contextual representations.
On one hand, GATE enjoys the capability of capturing long-range dependencies, which is crucial for languages with longer sentences, e.g., Arabic.\footnote{After tokenization, on average, ACE 2005 English and Arabic sentences have approximately 30 and 210 words, respectively.}
On the other hand, GATE is agnostic to language word order as it uses syntactic distance to model pairwise relationship between words.
This characteristic makes GATE suitable to transfer across typologically diverse languages, e.g., English to Arabic.
One crucial property of GATE is that it allows information propagation among different heads in the multi-head attention structure based on syntactic distances, which allows to learn the correlation between different mention types and target labels.

We conduct experiments on cross-lingual transfer among English, Chinese, and Arabic languages using the ACE 2005 benchmark \cite{walker2006ace}.
The experimental results demonstrate that GATE outperforms three recently proposed relation and event extraction methods by a significant margin.\footnote{Code available at https://github.com/wasiahmad/GATE}
We perform a thorough ablation study and analysis, which shows that GATE is less sensitive to source language's characteristics (e.g., word order, sentence structure) and thus excels in the cross-lingual transfer.

\section{Task Description}

In this paper, we focus on \emph{sentence-level} relation extraction \cite{subburathinam2019cross, florian2019neural} and event extraction \cite{subburathinam2019cross, liu2019neural} tasks.
Below, we first introduce the basic concepts, the notations, as well as define the problem and the scope of the work. 


\smallskip
\noindent\textbf{Relation Extraction} is the task of identifying the relation type of an ordered pair of entity mentions.
Formally, given a pair of entity mentions from a sentence $s$ - $(e_{s}, e_{o}; s)$ where $e_{s}$ and $e_{o}$ denoted as the subject and object entities respectively, the relation extraction (RE) task is defined as predicting the relation $r \in R \cup \{{\tt None}\}$ between the entity mentions, where $R$ is a pre-defined set of relation types.
In the example provided in Figure \ref{fig:example}, there is a {\tt PHYS:Located} relation between the entity mentions ``Terrorists'' and ``hotel''.

\smallskip
\noindent\textbf{Event Extraction} can be decomposed into two sub-tasks, \emph{Event Detection} and \emph{Event Argument Role Labeling}. 
Event detection refers to the task of identifying \emph{event triggers} (the words or phrases that express event occurrences) and their types.
In the example shown in Figure \ref{fig:example}, the word ``firing'' triggers the {\tt Attack} event.

Event argument role labeling (EARL) is defined as predicting whether words or phrases (arguments) participate in events and their roles. 
Formally, given an event trigger $e_{t}$ and a mention $e_{a}$ (an entity, time expression, or value) from a sentence $s$, the argument role labeling refers to predicting the mention's role $r \in R \cup \{{\tt None}\}$, where $R$ is a pre-defined set of role labels.
In Figure \ref{fig:example}, the ``Terrorists'' and ``hotel'' entities are the arguments of the {\tt Attack} event and they have the {\tt Attacker} and {\tt Place} role labels, respectively.

In this work, we focus on the EARL task; we assume event mentions (triggers) of the input sentence are provided.


\smallskip
\noindent\textbf{Zero-Short Cross-Lingual Transfer} refers to the setting, where there is no labeled examples available for the \emph{target} language.
We train neural relation extraction and event argument role labeling models on one (single-source) or multiple (multi-source) \emph{source} languages and then deploy the models in target languages. 
The overall cross-lingual transfer approach consists of four steps:
\begin{enumerate}
    \item Convert the input sentence into a language-universal tree structure using an off-the-shelf universal dependency parser, e.g., {UDP}ipe\footnote{http://ufal.mff.cuni.cz/udpipe} \cite{straka-strakova-2017-tokenizing}.
    \item Embed the words in the sentence into a shared semantic space across languages. We use off-the-shelf multilingual contextual encoders \cite{devlin-etal-2019-bert, conneau2019unsupervised} to form the word representations. To enrich the word representations, we concatenate them with \emph{universal}
    part-of-speech (POS) tag, dependency relation, and entity type embeddings \cite{subburathinam2019cross}.
    We collectively refer them as \emph{language-universal} features.
    \item Based on the word representations, we encode the input sentence using the proposed GATE architecture that leverages the syntactic depth and distance information. Note that this step is the main focus of this work. 
    \item A pair of classifier predicts the target relation and argument role labels based on the encoded representations.
\end{enumerate}


\section{Approach}
Our proposed approach GATE revises the multi-head attention architecture in Transformer Encoder \cite{vaswani2017attention} to model syntactic information while encoding a sequence of input vectors (represent the words in a sentence) into contextualized representations.
We first review the standard multi-head attention mechanism 
(\S \ref{sec:trans_enc}).
Then, we introduce our proposed method GATE (\S \ref{sec:gate}).
Finally, we describe how we perform relation extraction (\S \ref{sec:rel_ext}) and event argument role labeling (\S \ref{sec:eve_ext}) tasks.

\subsection{Transformer Encoder} 
\label{sec:trans_enc}

Unlike recent works \cite{zhang-etal-2018-graph, subburathinam2019cross} that use GCNs \cite{kipf2017semi} to encode the input sequences into contextualized representations, we propose to employ Transformer encoder as it excels in capturing long-range dependencies.
First, the sequence of input word vectors, $x = [x_1, \ldots, x_{|x|}]$ where $x_i \in R^d$ are packed into a matrix $H^0 = [x_1, \ldots, x_{|x|}]$.
Then an $L$-layer Transformer Encoder $H^l = Transformer_{l} (H^{l-1})$, $l \in [1, L]$ takes $H^0$ as input and generates different levels of latent representations $H^l = [h_1^l, \ldots, h_{|x|}^l]$, recursively.
Typically the latent representations generated by the last layer ($L$-th layer) are used as the contextual representations of the input words.
To aggregate the output vectors of the previous layer, multiple ($n_h$) self-attention heads are employed in each Transformer layer. 
For the $l$-th Transformer layer, the output of the previous layer $H^{l-1} \in \mathbb{R}^{|x| \times d_{model}}$ is first linearly projected to queries $Q$, keys $K$, and values $V$ using parameter matrices $W^Q_l, W^K_l \in \mathbb{R}^{d_{model} \times d_k}$ and $W^V_l \in \mathbb{R}^{d_{model} \times d_v}$, respectively.
\begin{linenomath*}
\begin{gather*}
    Q_l = H^{l-1} W^Q_l, K_l = H^{l-1} W^K_l, V_l = H^{l-1} W^V_l.
\end{gather*}
\end{linenomath*}
The output of a self-attention head $A_l$ is computed as:
\begin{linenomath*}
\begin{equation}
\label{eq:attention}
    A_l = \softmax\left(\frac{QK^T}{\sqrt{d_k}} + M\right) V_l,
\end{equation}
\end{linenomath*}
where the matrix $M \in \mathbb{R}^{|x|\times|x|}$ determines whether a pair of tokens can attend each other. 
\begin{linenomath*}
\begin{align}
\label{eq:mask}
    M_{ij} = \begin{cases}
                0, & \text{allow to attend} \\
                -\infty, & \text{prevent from attending}
             \end{cases}
\end{align}
\end{linenomath*}

The matrix $M$ is deduced as a \emph{mask}.
By default, the matrix M is a \emph{zero-matrix}.
In the next section, we discuss how we manipulate the mask matrix $M$ to incorporate syntactic depth and distance information in sentence representations.


\begin{figure}[t]
\centering
\includegraphics[width=0.73\linewidth]{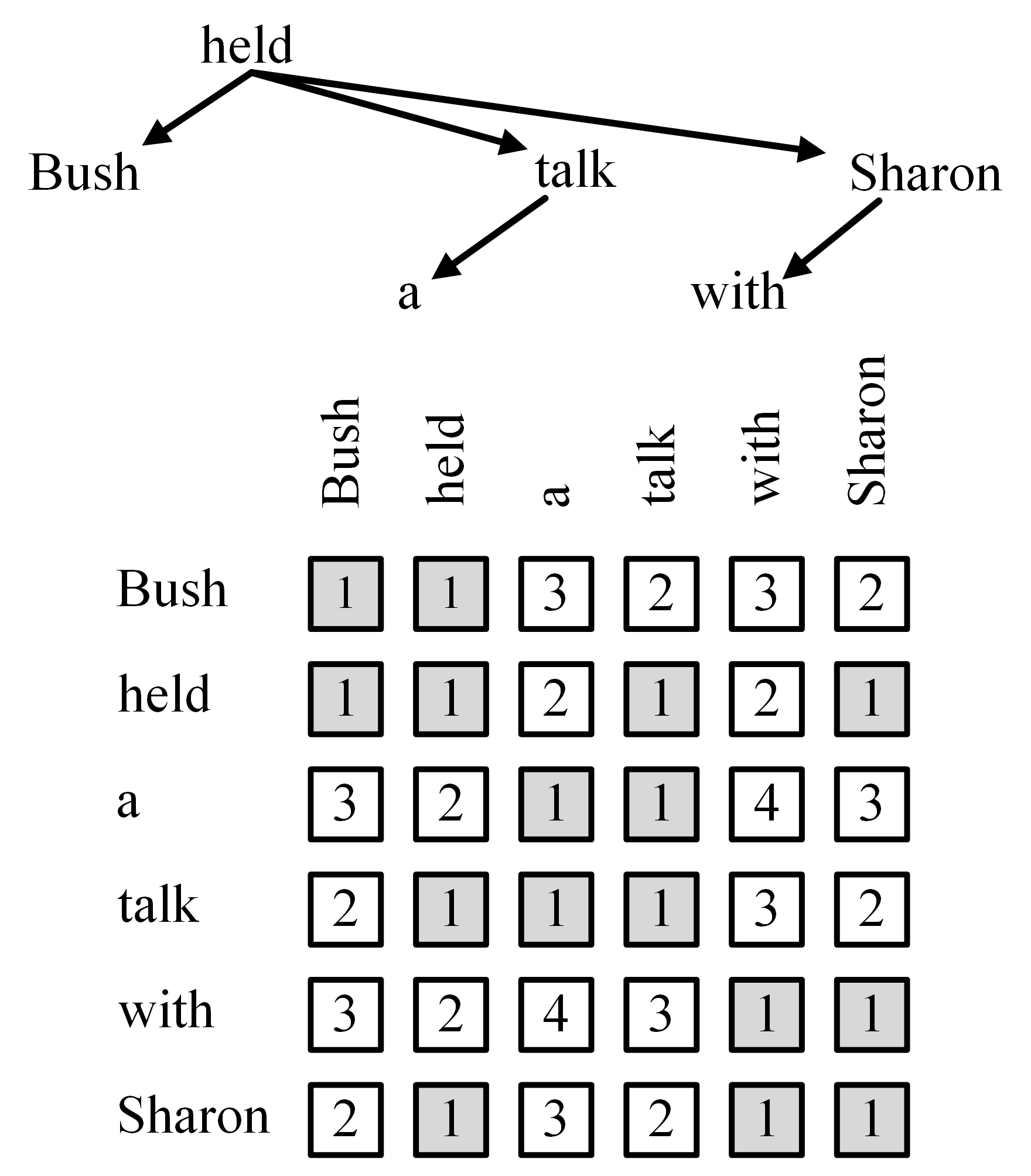}
\caption{
Distance matrix showing the shortest path distances between all pairs of words.
The dependency arc direction is ignored while computing pairwise distances.
The diagonal value is set to $1$, indicating a self-loop. If we set the values in white cells (with value $> 1$) to $0$, the distance matrix becomes an adjacency matrix.
}
\label{fig:distance_mask}
\vspace{-2mm}
\end{figure}

\subsection{Graph Attention Transformer Encoder}
\label{sec:gate}

The self-attention as described in \S \ref{sec:trans_enc} learns how much attention to put on words in a text sequence when encoding a word at a given position.
In this work, we revise the self-attention mechanism such that it takes into account the syntactic structure and distances when a token attends to all the other tokens.
The key idea is to manipulate the mask matrix to impose the graph structure and retrofit the attention weights based on pairwise syntactic distances. 
We use the universal dependency parse of a sentence and compute the syntactic (shortest path) distances between every pair of words.
We illustrate an example in Figure \ref{fig:distance_mask}.

We denote distance matrix $D \in \mathbb{R}^{|x| \times |x|}$ where $D_{ij}$ represents the syntactic distance between words at position $i$ and $j$ in the input sequence.
If we want to allow tokens to attend their adjacent tokens (that are $1$ hop away) at each layer, then we can set the mask matrix as follows.
\begin{linenomath*}
\begin{align*}
    M_{ij} = \begin{cases}
                0, & D_{ij} = 1 \\
                -\infty, & \text{otherwise}
             \end{cases}
\end{align*}
\end{linenomath*}
We generalize this notion to model a distance based attention; allowing tokens to attend tokens that are within distance $\delta$ (hyper-parameter).
\begin{linenomath*}
\begin{align}
\label{eq:distance_mask}
    M_{ij} = \begin{cases}
                0, & D_{ij} \leq \delta \\
                -\infty, & \text{otherwise}
             \end{cases}
\end{align}
\end{linenomath*}


During our preliminary analysis, we observed that syntactic distances between entity mentions or event mentions often correlate with the target label.
For example, if an {\tt ORG} entity mention appears closer to a {\tt PER} entity than a {\tt LOC} entity, then the \{{\tt PER, ORG}\} entity pair is more likely to have the {\tt PHYS:Located} relation.
We hypothesize that modeling syntactic distance between words can help to identify complex semantic structure such as events and entity relations.
Hence we revise the attention head $A_l$ (defined in Eq. \eqref{eq:attention}) computation as follows.
\begin{linenomath*}
\begin{gather}
\label{eq:rev_attn}
    A_l = F\left(\softmax\left(\frac{QK^T}{\sqrt{d_k}} + M\right)\right) V_l.
\end{gather}
\end{linenomath*}
Here, $\softmax$ produces an attention matrix $P \in R^{|x|\times|x|}$ where $P_i$ denotes the attentions that $i$-th token pays to the all the tokens in the sentence, and $F$ is a function that modifies those attention weights.
We can treat $F$ as a parameterized function that can be learned based on distances.
However, we adopt a simple formulation of $F$ such that GATE pays more attention to tokens that are closer and less attention to tokens that are faraway in the parse tree.
We define the $(i,j)$-th element of the attention matrix produced by $F$ as:
\begin{linenomath*}
\begin{gather}
\label{eq:dist_function}
    F(P)_{ij} = \frac{P_{ij}}{Z_i D_{ij}},
\end{gather}
\end{linenomath*}
where  $Z_i = \sum_j \frac{P_{ij}}{D_{ij}}$ is the normalization factor and $D_{ij}$ is the distance between $i$-th and $j$-th token. 
We found this formulation of $F$ effective for the IE tasks.


\subsection{Relation Extractor}
\label{sec:rel_ext}
Relation Extractor predicts the relationship label (or {\tt None}) for each mention pair in a sentence.
For an input sentence $s$, GATE produces contextualized word representations $h_1^l, \ldots, h_{|x|}^l$ where $h_i^l \in \mathbb{R}^{d_{model}}$.
As different sentences and entity mentions may have different lengths, we perform max-pooling over their contextual representations to obtain fixed-length vectors.
Suppose for a pair of entity mentions $e_{s} = [h^l_{bs}, \ldots, h^l_{es}]$ and $e_{o} = [h^l_{bo}, \ldots, h^l_{eo}]$, we obtain single vector representations $\hat{e}_s$ and $\hat{e}_o$ by performing max-pooling.
Following \citet{zhang-etal-2018-graph, subburathinam2019cross}, we also obtain a vector representation for the sentence, $\hat{s}$ by applying max-pooling over $[h_1^l, \ldots, h_{|x|}^l]$ and concatenate the three vectors.
Then the concatenation of the three vectors $[\hat{e}_s;\hat{e}_o;\hat{s}]$ are fed to a linear classifier followed by a Softmax layer to predict the relation type between entity mentions $e_s$ and $e_o$ as follows.
\begin{equation*}
    \mathcal{O}_r = \softmax(\bm{W}_r^T [\hat{e_s}; \hat{e_o}; \hat{s}] + \bm{b}_r),
\end{equation*}
where $\bm{W}_r \in R^{3d_{model} \times r}$ and $\bm{b}_r \in R^r$ are parameters, and $r$ is the total number of relation types.
The probability of $t$-th relation type is denoted as $P(r_t|s, e_s, e_o)$, which corresponds to the $t$-th element of $\mathcal{O}_r$.
To train the relation extractor, we adopt the cross-entropy loss.
\begin{equation*}
    \mathcal{L}_r = - \sum_{s=1}^N \sum_{o=1}^N \ \log(P(y_{so}^r|s, e_s, e_o)),
\end{equation*}
where $N$ is the number of entity mentions in the input sentence $s$ and $y^r_{so}$ denotes the ground truth relation type between entity mentions $e_s$ and $e_o$.

\subsection{Event Argument Role Labeler}
\label{sec:eve_ext}
Event argument role labeler predicts the argument mentions (or {\tt None} for non-argument mentions) of an event mention and assigns a role label to each argument from a pre-defined set of labels.
To label an argument candidate $e_a = [h_{ba}^l, \ldots, h_{ea}^l]$ for an event trigger $e_t = [h_{bt}^l, \ldots, h_{et}^l]$ in sentence $s = [h_1^l, \ldots, h_{|x|}^l]$, we apply max-pooling 
to form vectors $\hat{e}_a$, $\hat{e}_t$, and $\hat{s}$ respectively, which is same as that for relation extraction.
Then we concatenate the vectors ($[\hat{e_t}; \hat{e_a}; \hat{s}]$) and pass it through a linear classifier and Softmax layer to predict the role label as follows. 
\begin{equation*}
    \mathcal{O}_a = \softmax(\bm{W}_a^T [\hat{e_t}; \hat{e_a}; \hat{s}] + \bm{b}_a),
\end{equation*}
where $\bm{W}_a \in R^{3d_{model} \times r}$ and $\bm{b}_a \in R^r$ are parameters, and $r$ is the total number of argument role label types.
We optimize the role labeler by minimizing the cross-entropy loss.

\section{Experiment Setup}


\begin{table}[t]
\centering
\resizebox{\linewidth}{!}{%
\begin{tabular}{l@{\hskip 0.05in}|c@{\hskip 0.1in} c@{\hskip 0.1in} c@{\hskip 0.05in}|c@{\hskip 0.1in} c@{\hskip 0.1in} c@{\hskip 0.05in}}
\hline
& \multicolumn{3}{c|}{Sequential} & \multicolumn{3}{c}{Syntactic} \\ 
\cline{2-7}
& En & Zh & Ar & En & Zh & Ar \\ 
\hline
Relation Mention & 4.8 & 3.9 & 25.8 & 2.2 & 2.6 & 5.1  \\
Event Mention \& Arg. & 9.8 & 21.7 & 58.1 & 3.1 & 4.6 & 12.3 \\
\hline
\end{tabular}
}
\vspace{-2mm}
\caption{
Average sequential and syntactic (shortest path) distance between relation mentions and event mentions and their candidate arguments in ACE05 dataset.
Distances are computed by ignoring the order of mentions.
}
\label{table:dist_stat}
\vspace{-2mm}
\end{table}

\subsubsection{Dataset}
We conduct experiments based on the Automatic Content Extraction (ACE) 2005 corpus \cite{walker2006ace} that includes manual annotation of relation and event mentions (with their arguments) in three languages: English (En), Chinese (Zh), and Arabic (Ar).
We present the data statistics in Appendix.
ACE defines an ontology that includes 7 entity types, 18 relation subtypes, and 33 event subtypes.
We add a class label {\tt None} to denote that two entity mentions or a pair of an event mention and an argument candidate under consideration do not have a relationship belong to the target ontology.
We use the same dataset split as \citet{subburathinam2019cross} and follow their preprocessing steps.
We refer the readers to \citet{subburathinam2019cross} for the dataset preprocessing details.

\subsubsection{Evaluation Criteria}
Following the previous works \cite{ji-grishman-2008-refining, li-etal-2013-joint, li-ji-2014-incremental, subburathinam2019cross}, we set the evaluation criteria as, (1) a relation mention is correct if its predicted type and the head offsets of the two associated entity mentions are correct, and (2) an event argument role label is correct if the event type, offsets, and argument role label match any of the reference argument mentions.

\begin{table*}[t]
\centering
\begin{tabular}{l|c c c c c c|c c c c c c}
\hline
\multirow{4}{*}{Model} & \multicolumn{6}{c|}{Event Argument Role Labeling} & \multicolumn{6}{c}{Relation Extraction} \\
\cline{2-7}\cline{8-13}
& En & En & Zh & Zh & Ar & Ar & En & En & Zh & Zh & Ar & Ar \\ 
& $\Downarrow$ & $\Downarrow$ & $\Downarrow$ & $\Downarrow$ & $\Downarrow$ & $\Downarrow$ & $\Downarrow$ & $\Downarrow$ & $\Downarrow$ & $\Downarrow$ & $\Downarrow$ & $\Downarrow$ \\
& Zh & Ar & En & Ar & En & Zh & Zh & Ar & En & Ar & En & Zh \\ 
\hline
\multirow{1}{*}{\bone} & 41.8 & 55.6 & 41.2 & 52.9 & 39.6 & 40.8 & 56.7 & 65.3 & 65.9 & 59.7 & 59.6 & 46.3 \\
\multirow{1}{*}{\btwo} & 51.9 & 50.4 & 53.7 & 51.5 & 50.3 & 51.9 & 49.4 & 58.3 & 65.0 & 55.0 & 56.7 & 42.4 \\
\multirow{1}{*}{\bthree} & 60.4 & 53.9 & 55.7 & 52.5 & 50.7 & 50.9 & 53.7 & 63.9 & 70.9 & 57.6 & 67.1 & 55.7 \\
\multirow{1}{*}{\bfour} & 61.5 & 55.0 & 58.0 & 57.7 & 54.3 & 57.0 & 57.1 & 63.4 & 69.6 & 60.6 & 67.0 & 52.6 \\
\multirow{1}{*}{\bfive} & 62.3 & 60.8 & 57.3 & 66.3 & \textbf{57.5} & \textbf{59.8} & \textbf{58.0} & 59.9 & 70.0 & 55.6 & 66.5 & \textbf{56.5} \\
\multirow{1}{*}{GATE (this work)} & \textbf{63.2} & \textbf{68.5} & \textbf{59.3} & \textbf{69.2} & 53.9 & 57.8 & {55.1} & \textbf{66.8} & \textbf{71.5} & \textbf{61.2} & \textbf{69.0} & 54.3\\
\hline
\end{tabular}
\vspace{-2mm}
\caption{
Single-source transfer results (F-score \% on the test set) using perfect event triggers and entity mentions.
The language on top and bottom of $\Downarrow$ denotes the source and target languages, respectively.
}
\label{table:sst_results}
\vspace{-2mm}
\end{table*}


\begin{table}[t]
\centering
\vspace{2mm}
\begin{tabular}{l|c c c}
\hline
\multirow{3}{*}{Model} & \{En, Zh\} & \{En, Ar\} & \{Zh, Ar\} \\
& $\Downarrow$ & $\Downarrow$ & $\Downarrow$ \\
& Ar & Zh & En \\ 
\hline
\multicolumn{4}{l}{{Event Argument Role Labeling}} \\
\hline
\bone & 57.0 & 44.5 & 44.8 \\
\btwo & 58.9 & 56.2 & 57.9 \\
\bthree & 53.5 & 62.5 & 60.8 \\
\bfour & 59.5 & 62.0 & 60.7 \\
\bfive & 71.1 & \textbf{68.4} & \textbf{62.2} \\
GATE (this work) & \textbf{73.9} & 65.3 & 61.3 \\
\hline
\multicolumn{4}{l}{{Relation Extraction}} \\
\hline
\bone & 66.8 & 54.4 & 69.5 \\
\btwo & 64.0 & 46.6 & 65.8 \\
\bthree & 66.5 & 60.5 & 73.0 \\
\bfour & \textbf{68.3} & 59.3 & 73.7 \\
\bfive & 65.0 & \textbf{62.3} & 73.8 \\
GATE (this work) & 67.0 & 57.9 & \textbf{74.1} \\
\hline
\end{tabular}
\vspace{-2mm}
\caption{
Multi-source transfer results (F-score \% on the test set) using perfect event triggers and entity mentions.
The language on top and bottom of $\Downarrow$ denotes the source and target languages, respectively.
}
\label{table:mst_results}
\vspace{-2mm}
\end{table}

\subsubsection{Baseline Models}
To compare GATE on relation and event argument role labeling tasks, we chose three recently proposed approaches as baselines. The source code of the baselines are not publicly available at the time this research is conducted. Therefore, we reimplemented them.

\smallskip\noindent
{$\bullet$ \bone} \cite{liu2019neural} is a context-dependent lexical mapping approach where each word in a source language sentence is mapped to its best-suited translation in the target language.
We use multilingual word embeddings \cite{joulin-etal-2018-loss} as the continuous representations of tokens along with the language-universal features embeddings including part-of-speech (POS) tag embedding, dependency relation label embedding, and entity type embedding.\footnote{Due to the design principle of \citet{liu2019neural}, we cannot use multilingual contextual encoders in {\bone}.}
Since this model focuses on the target language, we train this baseline for each combination of source and target languages.

\smallskip\noindent
{$\bullet$ \btwo} \cite{subburathinam2019cross}
uses GCN \cite{kipf2017semi} to learn structured common space representation.
To embed the tokens in an input sentence, we use multilingual contextual representations \cite{devlin-etal-2019-bert, conneau2019unsupervised} and the language-universal feature embeddings.
We train this baseline on the source languages and directly evaluate on the target languages.

\smallskip\noindent
{$\bullet$ \bthree} \cite{florian2019neural} uses a bidirectional Long Short-Term Memory (LSTM) type recurrnet neural networks \cite{hochreiter1997long} to learn contextual representation.
We feed language-universal features for words in a sentence, constructed in the same way as \citet{subburathinam2019cross}.
We train and evaluate this baseline in the same way as {\btwo}.

In addition to the above three baseline methods, we compare GATE with the following two encoding methods.

\smallskip\noindent
{$\bullet$ \bfour} \cite{vaswani2017attention} uses multi-head self-attention mechanism and is the base structure of our proposed model, GATE.
Note that GATE has the same number of parameters as {\bfour} since GATE does not introduce any new parameter while modeling the pairwise syntactic distance into the self-attention mechanism. Therefore, we credit the GATE's improvements over the Transformer to its distance-based attention modeling strategy.

\smallskip\noindent
{$\bullet$ \bfive} \cite{shaw-etal-2018-self} uses relative position representations to encode the structure of the input sequences.
This method uses the pairwise \emph{sequential} distances while GATE uses pairwise \emph{syntactic} distances to model attentions between tokens.

\subsubsection{Implementation Details}
To embed words into vector representations, we use multilingual BERT (M-BERT) \cite{devlin-etal-2019-bert}.
Note that we do not fine-tune M-BERT, but only use it as a feature extractor.
We use the universal part-of-speech (POS) tags, dependency relation labels, and seven entity types defined by ACE: person, organization, geo-Political entity, location, facility, weapon, and vehicle.
We embed these language-universal features into fixed-length vectors and concatenate them with M-BERT vectors to form the input word representations.
We set the model size ($d_{model}$), number of encoder layers ($L$), and attention heads ($n_h$) in multi-head to 512, 1, and 8 respectively.
We tune the distance threshold $\delta$ (as shown in Eq. \eqref{eq:distance_mask}) in $[1, 2, 4, 8, \infty]$ for each attention head on each source language (more details are provided in the supplementary).

We implement all the baselines and our approach based on the implementation of \citet{zhang-etal-2018-graph}
and OpenNMT \cite{klein2017opennmt}. 
We used {\tt transformers}\footnote{https://github.com/huggingface/transformers}
to extract M-BERT and XLM-R features.
We provide a detailed description of the dataset, hyper-parameters, and training of the baselines and our approach in the supplementary.

\section{Results and Analysis}
We compare GATE with five baseline approaches on event argument role labeling (EARL) and relation extraction (RE) tasks, and the results are presented in Table \ref{table:sst_results} and \ref{table:mst_results}.

\subsubsection{Single-source transfer}
In the single-source transfer setting, all the models are individually trained on \emph{one} source language, e.g., English and directly evaluated on the other two languages (target), e.g., Chinese and Arabic.
Table \ref{table:sst_results} shows that GATE outperforms all the baselines in four out of six transfer directions on both tasks.
{\bthree} surprisingly outperforms {\btwo} in most settings, although {\bthree} uses a BiLSTM that is not suitable to transfer across syntactically different languages \cite{ahmad-etal-2019-difficulties}.
We hypothesize the reason being GCNs cannot capture long-range dependencies, which is crucial for the two tasks.
In comparison, by modeling distance-based pairwise relationships among words, GATE excels in cross-lingual transfer.

A comparison between {\bfour} and GATE demonstrates the effectiveness of syntactic distance-based self-attention over the standard mechanism.
From Table \ref{table:sst_results}, we see GATE outperforms {\bfour} with an average improvement of 4.7\% and 1.3\% in EARL and RE tasks, respectively.
Due to implicitly modeling graph structure, {\bfive} performs effectively.
However, GATE achieves an average improvement of 1.3\% and 1.9\% in EARL and RE tasks over {\bfive}.
Overall, the significant performance improvements achieved by GATE corroborate our hypothesis that syntactic distance-based attention helps in the cross-lingual transfer.

\subsubsection{Multi-source transfer}
In the multi-source cross-lingual transfer, the models are trained on a pair of languages: \{English, Chinese\}, \{English, Arabic\}, and \{Chinese, Arabic\}.
Hence, the models observe more examples during training, and as a result, the cross-lingual transfer performance improves compared to the single-source transfer setting.
In Table \ref{table:mst_results}, we see GATE outperforms the previous three IE approaches in multi-source transfer settings, except on RE for the source:\{English, Arabic\} and target: Chinese language setting.
On the other hand, GATE performs competitively to  {\bfour} and {\bfive} baselines.
Due to observing more training examples, {\bfour} and {\bfive} perform more effectively in this setting. 
The overall result indicates that GATE more efficiently learns transferable representations for the IE tasks.


\begin{table}[!t]
\centering
\begin{tabular}{l|c c|c c}
\hline
\multirow{2}{*}{Model} & \multicolumn{2}{c|}{EARL} & \multicolumn{2}{c}{RE} \\
\cline{2-5}
& Chinese & Arabic & Chinese & Arabic \\
\hline
\multicolumn{5}{l}{\citet{wang-etal-2019-self}} \\
\hline
Absolute  & 61.2 & 53.5 & 57.8 & 65.2 \\
Relative  & 55.3 & 47.1 & \textbf{58.1} & 66.4 \\
\hline
GATE      & \textbf{63.2} & \textbf{68.5} & 55.1 & \textbf{66.8} \\
\hline
\end{tabular}
\vspace{-2mm}
\caption{
GATE vs. \citet{wang-etal-2019-self} results (F-score \%) on event argument role labeling (EARL) and relation extraction (RE); using English as source and Chinese, Arabic as the target languages, respectively.
To limit the maximum relative position, the clipping distance is set to 10 and 5 for EARL and RE tasks, respectively.
}
\label{ablation:position}
\vspace{-2mm}
\end{table}


\subsubsection{Encoding dependency structure}
\label{subsec:position_abl}
GATE encodes the dependency structure of sentences by guiding the attention mechanism in self-attention networks (SANs).
However, an alternative way to encode the sentence structure is through positional encoding for SANs.
Conceptually, the key difference is the modeling of syntactic distances to capture fine-grained relations among tokens.
Hence, we compare these two notions of encoding the dependency structure to emphasize the promise of modeling syntactic distances.

To this end, we compare the GATE with \citet{wang-etal-2019-self} that proposed structural position encoding using the dependency structure of sentences.
Results are presented in Table \ref{ablation:position}.
We see that \citet{wang-etal-2019-self} performs well on RE but poorly  on EARL, especially on the Arabic language.
While GATE directly uses syntactic distances between tokens to guide the self-attention mechanism, \citet{wang-etal-2019-self} learns parameters to encode structural positions that can become sensitive to the source language.
For example, the average shortest path distance between event mentions and their candidate arguments in English and Arabic is 3.1 and 12.3, respectively (see Table \ref{table:dist_stat}).
As a result, a model trained in English may learn only to attend closer tokens, thus fails to generalize on Arabic. 

Moreover, we anticipate that different order of subject and verb in English and Arabic\footnote{According to WALS \cite{dryer2013world}, the order of subject (S), object (O), and verb (V) for English, Chinese and Arabic is SVO, SVO, and VSO.} causes \citet{wang-etal-2019-self} to transfer poorly on the EARL task (as event triggers are mostly verbs).
To verify our anticipation, we modify the relative structural position encoding \cite{wang-etal-2019-self} by dropping the directional information \cite{ahmad-etal-2019-difficulties}, and observed a  performance increase from 47.1 to 52.2 for English to Arabic language transfer.
In comparison, GATE is order-agnostic as it models syntactic distance; hence, it has a better transferability across typologically diverse languages. 


\begin{table}[t]
\centering
\begin{tabular}{l|c c|c c}
\hline
\multirow{2}{*}{Model} & \multicolumn{2}{c|}{EARL} & \multicolumn{2}{c}{RE} \\
\cline{2-5}
& English & Chinese$^\ast$ & English & Chinese$^\ast$ \\
\hline
\btwo & 51.5 & 56.3 & 46.9 & 50.7 \\
\bthree & 55.6 & 59.3 & 56.8 & \textbf{62.0} \\
GATE & \textbf{63.8} & \textbf{64.2} & \textbf{58.8} & 57.0 \\
\hline
\end{tabular}
\vspace{-2mm}
\caption{
Event argument role labeling (EARL) and relation extraction (RE) results (F-score \%); using Chinese as the source and English as the target language.
$^\ast$ indicates the English examples are translated into Chinese using {\tt Google Cloud Translate}.
}
\label{table:gtranslate}
\vspace{-4mm}
\end{table}

\begin{figure}[t]
\captionsetup[subfigure]{labelformat=empty}
\centering
\hspace{-4mm}
\subfloat[\label{fig2:sub1}]
{
\includegraphics[width=1.0\linewidth]{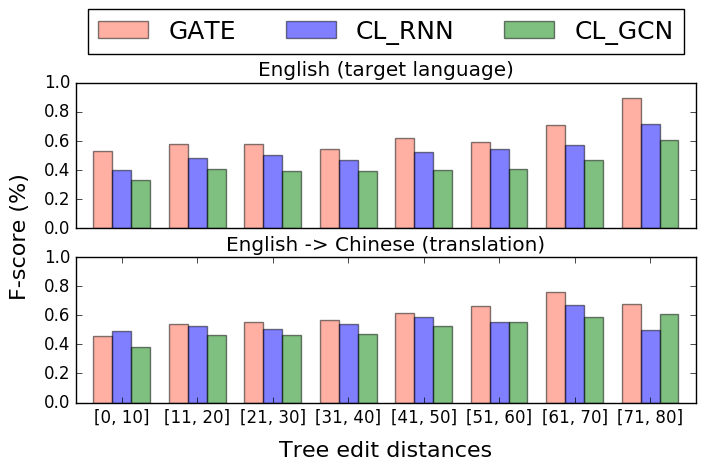}
}
\vspace{-7mm}
\caption{
Models trained on the Chinese language perform on event argument role labeling in English and their parallel Chinese sentences.
The parallel sentences have the same meaning but a different structure.
To quantify the structural difference between two parallel sentences, we compute the tree edit distances.
}
\vspace{-2mm}
\label{figure:gtranslate}
\end{figure}

\subsubsection{Sensitivity towards source language}
Intuitively, an RE or EARL model would transfer well on target languages if the model is less sensitive towards the source language characteristics (e.g., word order, grammar structure).
To measure sensitivity towards the source language, we evaluate the model performance on the target language and their parallel (translated) source language sentences. 
We hypothesize that if a model performs significantly well on the translated source language sentences, then the model is more sensitive towards the source language and may not be ideal for cross-lingual transfer.
To test the models on this hypothesis, we translate all the ACE05 English test set examples into Chinese using {\tt Google Cloud Translate}.\footnote{Details are provided in the supplementary.}
We train GATE and two baselines on the Chinese and evaluate them on both English (test set) examples and their Chinese translations.
To quantify the difference between the dependency structure of an English and its Chinese translation sentences, we compute \emph{edit distance} between two tree structures using the APTED\footnote{https://pypi.org/project/apted/} algorithm \cite{pawlik2015efficient, pawlik2016tree}.

The results are presented in Table \ref{table:gtranslate}.
We see that  {\btwo} and  {\bthree} have much higher accuracy on the translated (Chinese) sentences than the target language (English) sentences.
On the other hand, GATE makes a roughly similar number of correct predictions when the target and translated sentences are given as input.
Figure \ref{figure:gtranslate} illustrates how the models perform when the structural distance between target sentences and their translation increases. 
The results suggest that GATE performs substantially better than the baselines when the target language sentences are structurally different from the source language. 
The overall findings signal that GATE is less sensitive to source language features, and we credit this to the modeling of distance-based syntactic relationships between words.
We acknowledge that there might be other factors associated with a model's language sensitivity.
However, we leave the detailed analysis for measuring a model's sensitivity towards languages as future work.

\begin{table}[t]
\centering
\vspace{2mm}
\begin{tabular}{l|c c|c c}
\hline
\multirow{2}{*}{Word features} & \multicolumn{2}{c|}{EARL} & \multicolumn{2}{c}{RE} \\
\cline{2-5}
& Chinese & Arabic & Chinese & Arabic \\
\hline
Multi-WE & 35.9 & 43.7 & 41.0 & 54.9 \\
M-BERT & \textbf{57.1} & 54.8 & \textbf{55.1} & 66.8 \\
XLM-R & 51.8 & \textbf{61.7} & 51.4 & \textbf{68.1} \\
\hline
\end{tabular}
\vspace{-2mm}
\caption{
Contribution of multilingual word embeddings (Multi-WE) \cite{joulin-etal-2018-loss}, M-BERT \cite{devlin-etal-2019-bert}, and XLM-R \cite{conneau2019unsupervised} as a source of word features; using English as source and Chinese, Arabic as the target languages, respectively.
}
\label{ablation:word_rep}
\end{table}



\begin{table}[t]
\centering
\begin{tabular}{l|c c|c c}
\hline
\multirow{2}{*}{Input features} & \multicolumn{2}{c|}{EARL} & \multicolumn{2}{c}{RE} \\
\cline{2-5}
& Chinese & Arabic & Chinese & Arabic \\
\hline
M-BERT & 52.5 & 47.4 & 44.0 & 49.7 \\
\enspace + POS tag  & 49.3 & 47.5 & 44.1 & 47.0 \\
\enspace + Dep. label & 49.7 & 51.0 & 48.6 & 47.0 \\
\enspace + Entity type & \textbf{57.8} & \textbf{60.2} & \textbf{56.3} & \textbf{63.0} \\
\hline
\end{tabular}
\vspace{-2mm}
\caption{
Ablation on the use of language-universal features (part-of-speech (POS) tag, dependency relation label, and entity type) in GATE (F-score (\%); using English as source and Chinese, Arabic as the target languages, respectively.
}
\label{ablation:input_features}
\vspace{-2mm}
\end{table}


\subsubsection{Ablation study}
We perform a detailed ablation study on language-universal features and sources of word features to examine their individual impact on cross-lingual transfer.
The results are presented in Table \ref{ablation:word_rep} and \ref{ablation:input_features}.
We observed that M-BERT and XLM-R produced word features performed better in Chinese and Arabic, respectively, while they are comparable in English.
On average M-BERT performs better, and thus we chose it as the word feature extractor in all our experiments.
Table \ref{ablation:input_features} shows that part-of-speech and dependency relation embedding has a limited contribution. This is perhaps due to the tokenization errors, as pointed out by \citet{subburathinam2019cross}.
However, the use of language-universal features is useful, particularly when we have minimal training data.
We provide more analysis and results in the supplementary.

\section{Related Work}
Relation and event extraction has drawn significant attention from the natural language processing (NLP) community.
Most of the approaches developed in past several years are based on supervised machine learning, using either symbolic features \cite{ahn-2006-stages, ji-grishman-2008-refining, liao-grishman-2010-using, hong-etal-2011-using, li-etal-2013-joint, li-ji-2014-incremental} 
or distributional features \cite{liao-grishman-2011-acquiring, nguyen2016joint, miwa-bansal-2016-end, liu2018event, zhang2018using, liu2018similar, chen-etal-2015-event, nguyen2015event, zeng-etal-2014-relation, peng2017cross, nguyen2018graph, zhang-etal-2018-graph, subburathinam2019cross, liu2019neural, huang2020event} from a large number of annotations.
Joint learning or inference \cite{bekoulis-etal-2018-adversarial, li-etal-2014-constructing, zhang2019joint, liu-etal-2018-jointly, nguyen2016joint, yang-mitchell-2016-joint, han2019joint, han2020knowledge} are also among the noteworthy techniques.

Most previous works on cross-lingual transfer for relation and event extraction are based on annotation projection \cite{kim2010cross, kim-lee-2012-graph}, bilingual dictionaries \cite{hsi-etal-2016-leveraging, florian2019neural}, parallel data \cite{chen-ji-2009-one, kim-etal-2010-cross, qian-etal-2014-bilingual} or
machine translation \cite{zhu-etal-2014-bilingual, faruqui-kumar-2015-multilingual, zou-etal-2018-adversarial}.
Learning common patterns across languages is also explored \cite{lin-etal-2017-neural, wang-etal-2018-adversarial, liu2018event}.
In contrast to these approaches, \citet{subburathinam2019cross, liu2019neural} proposed to use graph convolutional networks (GCNs) \cite{kipf2017semi} to learn multi-lingual structured representations.
However, GCNs struggle to model long-range dependencies or disconnected words in the dependency tree.
To overcome the limitation, we use the syntactic distances to weigh the attentions while learning contextualized representations via the multi-head attention mechanism \cite{vaswani2017attention}.

Moreover, our proposed syntax driven distance-based attention modeling helps to mitigate the word order difference issue \cite{ahmad-etal-2019-difficulties} that hinders cross-lingual transfer. Prior works studied dependency structure modeling \cite{liu2019neural}, source reordering \cite{rasooli-collins-2019-low}, adversarial training \cite{ahmad-etal-2019-cross}, constrained inference \cite{meng-etal-2019-target} to tackle word order differences across typologically different languages.

\section{Conclusion}
In this paper, we proposed to model fine-grained syntactic structural information based on the dependency parse of a sentence.
We developed a {\bf G}raph {\bf A}ttention {\bf T}ransformer {\bf E}ncoder (GATE) to generate structured contextual representations.
Extensive experiments on three languages demonstrates the effectiveness of GATE in cross-lingual relation and event extraction. 
In the future, we want to explore other sources of language-universal information to improve structured representation learning.

\newpage
\section*{ Acknowledgments}
This work was supported in part by National Science Foundation (NSF) Grant OAC 1920462 and
the Intelligence Advanced Research Projects Activity (IARPA) via Contract No. 2019-19051600007.

\bibliography{aaai2021}

\begin{thebibliography}{67}
\providecommand{\natexlab}[1]{#1}
\providecommand{\url}[1]{\texttt{#1}}
\providecommand{\urlprefix}{URL }
\expandafter\ifx\csname urlstyle\endcsname\relax
  \providecommand{\doi}[1]{doi:\discretionary{}{}{}#1}\else
  \providecommand{\doi}{doi:\discretionary{}{}{}\begingroup
  \urlstyle{rm}\Url}\fi

\bibitem[{Ahmad et~al.(2019{\natexlab{a}})Ahmad, Zhang, Ma, Hovy, Chang, and
  Peng}]{ahmad-etal-2019-difficulties}
Ahmad, W.; Zhang, Z.; Ma, X.; Hovy, E.; Chang, K.-W.; and Peng, N.
  2019{\natexlab{a}}.
\newblock On Difficulties of Cross-Lingual Transfer with Order Differences: A
  Case Study on Dependency Parsing.
\newblock In \emph{Proceedings of NAACL}, 2440--2452.

\bibitem[{Ahmad et~al.(2019{\natexlab{b}})Ahmad, Zhang, Ma, Chang, and
  Peng}]{ahmad-etal-2019-cross}
Ahmad, W.~U.; Zhang, Z.; Ma, X.; Chang, K.-W.; and Peng, N. 2019{\natexlab{b}}.
\newblock Cross-Lingual Dependency Parsing with Unlabeled Auxiliary Languages.
\newblock In \emph{Proceedings of CoNLL}, 372--382.

\bibitem[{Ahn(2006)}]{ahn-2006-stages}
Ahn, D. 2006.
\newblock The stages of event extraction.
\newblock In \emph{Proceedings of the Workshop on Annotating and Reasoning
  about Time and Events}, 1--8.

\bibitem[{Bekoulis et~al.(2018)Bekoulis, Deleu, Demeester, and
  Develder}]{bekoulis-etal-2018-adversarial}
Bekoulis, G.; Deleu, J.; Demeester, T.; and Develder, C. 2018.
\newblock Adversarial training for multi-context joint entity and relation
  extraction.
\newblock In \emph{Proceedings of EMNLP}, 2830--2836.

\bibitem[{Bugliarello and Okazaki(2020)}]{bugliarello-okazaki-2020-enhancing}
Bugliarello, E.; and Okazaki, N. 2020.
\newblock Enhancing Machine Translation with Dependency-Aware Self-Attention.
\newblock In \emph{Proceedings of ACL}, 1618--1627.

\bibitem[{Chen et~al.(2015)Chen, Xu, Liu, Zeng, and
  Zhao}]{chen-etal-2015-event}
Chen, Y.; Xu, L.; Liu, K.; Zeng, D.; and Zhao, J. 2015.
\newblock Event Extraction via Dynamic Multi-Pooling Convolutional Neural
  Networks.
\newblock In \emph{Proceedings of ACL-IJCNLP}, 167--176.

\bibitem[{Chen and Ji(2009)}]{chen-ji-2009-one}
Chen, Z.; and Ji, H. 2009.
\newblock Can One Language Bootstrap the Other: A Case Study on Event
  Extraction.
\newblock In \emph{Proceedings of the {NAACL} {HLT} 2009 Workshop on
  Semi-supervised Learning for Natural Language Processing}, 66--74.

\bibitem[{Chen et~al.(2019)Chen, Chang, Chen, Nayak, and
  Ku}]{chen-etal-2019-uhop}
Chen, Z.-Y.; Chang, C.-H.; Chen, Y.-P.; Nayak, J.; and Ku, L.-W. 2019.
\newblock {UH}op: An Unrestricted-Hop Relation Extraction Framework for
  Knowledge-Based Question Answering.
\newblock In \emph{Proceedings of NAACL}, 345--356.

\bibitem[{Conneau et~al.(2020)Conneau, Khandelwal, Goyal, Chaudhary, Wenzek,
  Guzm{\'a}n, Grave, Ott, Zettlemoyer, and Stoyanov}]{conneau2019unsupervised}
Conneau, A.; Khandelwal, K.; Goyal, N.; Chaudhary, V.; Wenzek, G.; Guzm{\'a}n,
  F.; Grave, E.; Ott, M.; Zettlemoyer, L.; and Stoyanov, V. 2020.
\newblock Unsupervised Cross-lingual Representation Learning at Scale.
\newblock In \emph{Proceedings of ACL}.

\bibitem[{Deguchi, Tamura, and Ninomiya(2019)}]{deguchi-etal-2019-dependency}
Deguchi, H.; Tamura, A.; and Ninomiya, T. 2019.
\newblock Dependency-Based Self-Attention for Transformer {NMT}.
\newblock In \emph{Proceedings of RANLP}, 239--246.

\bibitem[{Devlin et~al.(2019)Devlin, Chang, Lee, and
  Toutanova}]{devlin-etal-2019-bert}
Devlin, J.; Chang, M.-W.; Lee, K.; and Toutanova, K. 2019.
\newblock {BERT}: Pre-training of Deep Bidirectional Transformers for Language
  Understanding.
\newblock In \emph{Proceedings of NAACL}.

\bibitem[{Dryer and Haspelmath(2013)}]{dryer2013world}
Dryer, M.~S.; and Haspelmath, M. 2013.
\newblock The world atlas of language structures online .

\bibitem[{Faruqui and Kumar(2015)}]{faruqui-kumar-2015-multilingual}
Faruqui, M.; and Kumar, S. 2015.
\newblock Multilingual Open Relation Extraction Using Cross-lingual Projection.
\newblock In \emph{Proceedings of NAACL}, 1351--1356.

\bibitem[{Fu, Li, and Ma(2019)}]{fu2019graphrel}
Fu, T.-J.; Li, P.-H.; and Ma, W.-Y. 2019.
\newblock {G}raph{R}el: Modeling Text as Relational Graphs for Joint Entity and
  Relation Extraction.
\newblock In \emph{Proceedings of ACL}, 1409--1418.

\bibitem[{Han, Ning, and Peng(2019)}]{han2019joint}
Han, R.; Ning, Q.; and Peng, N. 2019.
\newblock Joint Event and Temporal Relation Extraction with Shared
  Representations and Structured Prediction.
\newblock In \emph{Proceedings of EMNLP}, 434--444.

\bibitem[{Han, Zhou, and Peng(2020)}]{han2020knowledge}
Han, R.; Zhou, Y.; and Peng, N. 2020.
\newblock Domain Knowledge Empowered Structured Neural Net for End-to-End Event
  Temporal Relation Extraction.
\newblock In \emph{Proceedings of EMNLP}, 5717--5729.

\bibitem[{Hochreiter and Schmidhuber(1997)}]{hochreiter1997long}
Hochreiter, S.; and Schmidhuber, J. 1997.
\newblock Long short-term memory.
\newblock \emph{Neural computation} 9(8): 1735--1780.

\bibitem[{Hong et~al.(2011)Hong, Zhang, Ma, Yao, Zhou, and
  Zhu}]{hong-etal-2011-using}
Hong, Y.; Zhang, J.; Ma, B.; Yao, J.; Zhou, G.; and Zhu, Q. 2011.
\newblock Using Cross-Entity Inference to Improve Event Extraction.
\newblock In \emph{Proceedings of NAACL}, 1127--1136.

\bibitem[{Hsi et~al.(2016)Hsi, Yang, Carbonell, and
  Xu}]{hsi-etal-2016-leveraging}
Hsi, A.; Yang, Y.; Carbonell, J.; and Xu, R. 2016.
\newblock Leveraging Multilingual Training for Limited Resource Event
  Extraction.
\newblock In \emph{Proceedings of COLING}, 1201--1210.

\bibitem[{Huang, Yang, and Peng(2020)}]{huang2020event}
Huang, K.-H.; Yang, M.; and Peng, N. 2020.
\newblock Biomedical Event Extraction with Hierarchical Knowledge Graphs.
\newblock In \emph{Findings of ACL: EMNLP 2020}.

\bibitem[{Huang et~al.(2018)Huang, Ji, Cho, Dagan, Riedel, and
  Voss}]{huang-etal-2018-zero}
Huang, L.; Ji, H.; Cho, K.; Dagan, I.; Riedel, S.; and Voss, C. 2018.
\newblock Zero-Shot Transfer Learning for Event Extraction.
\newblock In \emph{Proceedings of ACL}, 2160--2170.

\bibitem[{Ji and Grishman(2008)}]{ji-grishman-2008-refining}
Ji, H.; and Grishman, R. 2008.
\newblock Refining Event Extraction through Cross-Document Inference.
\newblock In \emph{Proceedings of ACL}, 254--262.

\bibitem[{Joulin et~al.(2018)Joulin, Bojanowski, Mikolov, J{\'e}gou, and
  Grave}]{joulin-etal-2018-loss}
Joulin, A.; Bojanowski, P.; Mikolov, T.; J{\'e}gou, H.; and Grave, E. 2018.
\newblock Loss in Translation: Learning Bilingual Word Mapping with a Retrieval
  Criterion.
\newblock In \emph{Proceedings of EMNLP}, 2979--2984.

\bibitem[{Kim et~al.(2010{\natexlab{a}})Kim, Jeong, Lee, and
  Lee}]{kim2010cross}
Kim, S.; Jeong, M.; Lee, J.; and Lee, G.~G. 2010{\natexlab{a}}.
\newblock A cross-lingual annotation projection approach for relation
  detection.
\newblock In \emph{Proceedings of COLING}, 564--571.

\bibitem[{Kim et~al.(2010{\natexlab{b}})Kim, Jeong, Lee, and
  Lee}]{kim-etal-2010-cross}
Kim, S.; Jeong, M.; Lee, J.; and Lee, G.~G. 2010{\natexlab{b}}.
\newblock A Cross-lingual Annotation Projection Approach for Relation
  Detection.
\newblock In \emph{Proceedings of COLING}, 564--571.

\bibitem[{Kim and Lee(2012)}]{kim-lee-2012-graph}
Kim, S.; and Lee, G.~G. 2012.
\newblock A Graph-based Cross-lingual Projection Approach for Weakly Supervised
  Relation Extraction.
\newblock In \emph{Proceedings of ACL}, 48--53.

\bibitem[{Kipf and Welling(2017)}]{kipf2017semi}
Kipf, T.~N.; and Welling, M. 2017.
\newblock Semi-supervised classification with graph convolutional networks.
\newblock In \emph{ICLR}.

\bibitem[{Klein et~al.(2017)Klein, Kim, Deng, Senellart, and
  Rush}]{klein2017opennmt}
Klein, G.; Kim, Y.; Deng, Y.; Senellart, J.; and Rush, A. 2017.
\newblock {O}pen{NMT}: Open-Source Toolkit for Neural Machine Translation.
\newblock In \emph{Proceedings of {ACL} 2017, System Demo}.

\bibitem[{Li and Ji(2014)}]{li-ji-2014-incremental}
Li, Q.; and Ji, H. 2014.
\newblock Incremental Joint Extraction of Entity Mentions and Relations.
\newblock In \emph{Proceedings of ACL}, 402--412.

\bibitem[{Li et~al.(2014)Li, Ji, Hong, and Li}]{li-etal-2014-constructing}
Li, Q.; Ji, H.; Hong, Y.; and Li, S. 2014.
\newblock Constructing Information Networks Using One Single Model.
\newblock In \emph{Proceedings of {EMNLP}}, 1846--1851.

\bibitem[{Li, Ji, and Huang(2013)}]{li-etal-2013-joint}
Li, Q.; Ji, H.; and Huang, L. 2013.
\newblock Joint Event Extraction via Structured Prediction with Global
  Features.
\newblock In \emph{Proceedings of ACL}, 73--82.

\bibitem[{Liao and Grishman(2010)}]{liao-grishman-2010-using}
Liao, S.; and Grishman, R. 2010.
\newblock Using Document Level Cross-Event Inference to Improve Event
  Extraction.
\newblock In \emph{Proceedings of ACL}, 789--797.

\bibitem[{Liao and Grishman(2011)}]{liao-grishman-2011-acquiring}
Liao, S.; and Grishman, R. 2011.
\newblock Acquiring Topic Features to improve Event Extraction: in Pre-selected
  and Balanced Collections.
\newblock In \emph{Proceedings of RANLP}, 9--16.

\bibitem[{Lin, Liu, and Sun(2017)}]{lin-etal-2017-neural}
Lin, Y.; Liu, Z.; and Sun, M. 2017.
\newblock Neural Relation Extraction with Multi-lingual Attention.
\newblock In \emph{Proceedings of ACL}.

\bibitem[{Lin et~al.(2015)Lin, Liu, Sun, Liu, and Zhu}]{lin2015learning}
Lin, Y.; Liu, Z.; Sun, M.; Liu, Y.; and Zhu, X. 2015.
\newblock Learning Entity and Relation Embeddings for Knowledge Graph
  Completion.
\newblock In \emph{Proceedings of AAAI}, 2181–2187.

\bibitem[{Liu et~al.(2018)Liu, Chen, Liu, and Zhao}]{liu2018event}
Liu, J.; Chen, Y.; Liu, K.; and Zhao, J. 2018.
\newblock Event detection via gated multilingual attention mechanism.
\newblock In \emph{Proceedings of AAAI}, 4865–4872.

\bibitem[{Liu et~al.(2019)Liu, Chen, Liu, and Zhao}]{liu2019neural}
Liu, J.; Chen, Y.; Liu, K.; and Zhao, J. 2019.
\newblock Neural Cross-Lingual Event Detection with Minimal Parallel Resources.
\newblock In \emph{Proceedings of EMNLP-IJCNLP}, 738--748.

\bibitem[{Liu, Luo, and Huang(2018)}]{liu-etal-2018-jointly}
Liu, X.; Luo, Z.; and Huang, H. 2018.
\newblock Jointly Multiple Events Extraction via Attention-based Graph
  Information Aggregation.
\newblock In \emph{Proceedings of EMNLP}, 1247--1256.

\bibitem[{Lu and Nguyen(2018)}]{liu2018similar}
Lu, W.; and Nguyen, T.~H. 2018.
\newblock Similar but not the Same: Word Sense Disambiguation Improves Event
  Detection via Neural Representation Matching.
\newblock In \emph{Proceedings of EMNLP}, 4822--4828.

\bibitem[{Meng, Peng, and Chang(2019)}]{meng-etal-2019-target}
Meng, T.; Peng, N.; and Chang, K.-W. 2019.
\newblock Target Language-Aware Constrained Inference for Cross-lingual
  Dependency Parsing.
\newblock In \emph{Proceedings of EMNLP-IJCNLP}, 1117--1128.

\bibitem[{Miwa and Bansal(2016)}]{miwa-bansal-2016-end}
Miwa, M.; and Bansal, M. 2016.
\newblock End-to-End Relation Extraction using {LSTM}s on Sequences and Tree
  Structures.
\newblock In \emph{Proceedings of ACL}, 1105--1116.

\bibitem[{Nguyen, Cho, and Grishman(2016)}]{nguyen2016joint}
Nguyen, T.~H.; Cho, K.; and Grishman, R. 2016.
\newblock Joint Event Extraction via Recurrent Neural Networks.
\newblock In \emph{Proceedings of NAACL}, 300--309.

\bibitem[{Nguyen and Grishman(2015)}]{nguyen2015event}
Nguyen, T.~H.; and Grishman, R. 2015.
\newblock Event Detection and Domain Adaptation with Convolutional Neural
  Networks.
\newblock In \emph{Proceedings of EMNLP-IJCNLP}, 365--371.

\bibitem[{Nguyen and Grishman(2018)}]{nguyen2018graph}
Nguyen, T.~H.; and Grishman, R. 2018.
\newblock Graph convolutional networks with argument-aware pooling for event
  detection.
\newblock In \emph{Proceedings of AAAI}.

\bibitem[{Ni and Florian(2019)}]{florian2019neural}
Ni, J.; and Florian, R. 2019.
\newblock Neural Cross-Lingual Relation Extraction Based on Bilingual Word
  Embedding Mapping.
\newblock In \emph{Proceedings of EMNLP-IJCNLP}, 399--409.

\bibitem[{Pawlik and Augsten(2015)}]{pawlik2015efficient}
Pawlik, M.; and Augsten, N. 2015.
\newblock Efficient computation of the tree edit distance.
\newblock \emph{ACM Transactions on Database Systems (TODS)} 40(1): 1--40.

\bibitem[{Pawlik and Augsten(2016)}]{pawlik2016tree}
Pawlik, M.; and Augsten, N. 2016.
\newblock Tree edit distance: Robust and memory-efficient.
\newblock \emph{Information Systems} 157--173.

\bibitem[{Peng et~al.(2017)Peng, Poon, Quirk, Toutanova, and
  Yih}]{peng2017cross}
Peng, N.; Poon, H.; Quirk, C.; Toutanova, K.; and Yih, W.-t. 2017.
\newblock Cross-sentence N-ary Relation Extraction with Graph LSTMs.
\newblock \emph{Transactions of ACL} .

\bibitem[{Qian et~al.(2014)Qian, Hui, Hu, Zhou, and
  Zhu}]{qian-etal-2014-bilingual}
Qian, L.; Hui, H.; Hu, Y.; Zhou, G.; and Zhu, Q. 2014.
\newblock Bilingual Active Learning for Relation Classification via Pseudo
  Parallel Corpora.
\newblock In \emph{Proceedings of ACL}, 582--592.

\bibitem[{Rasooli and Collins(2019)}]{rasooli-collins-2019-low}
Rasooli, M.~S.; and Collins, M. 2019.
\newblock Low-Resource Syntactic Transfer with Unsupervised Source Reordering.
\newblock In \emph{Proceedings of NAACL}, 3845--3856.

\bibitem[{Shaw, Uszkoreit, and Vaswani(2018)}]{shaw-etal-2018-self}
Shaw, P.; Uszkoreit, J.; and Vaswani, A. 2018.
\newblock Self-Attention with Relative Position Representations.
\newblock In \emph{Proceedings of NAACL}, 464--468.

\bibitem[{Straka and Strakov{\'a}(2017)}]{straka-strakova-2017-tokenizing}
Straka, M.; and Strakov{\'a}, J. 2017.
\newblock Tokenizing, {POS} Tagging, Lemmatizing and Parsing {UD} 2.0 with
  {UDP}ipe.
\newblock In \emph{Proceedings of the {C}o{NLL} 2017 Shared Task: Multilingual
  Parsing from Raw Text to Universal Dependencies}, 88--99.

\bibitem[{Subburathinam et~al.(2019)Subburathinam, Lu, Ji, May, Chang, Sil, and
  Voss}]{subburathinam2019cross}
Subburathinam, A.; Lu, D.; Ji, H.; May, J.; Chang, S.-F.; Sil, A.; and Voss, C.
  2019.
\newblock Cross-lingual Structure Transfer for Relation and Event Extraction.
\newblock In \emph{Proceedings of EMNLP-IJCNLP}, 313--325.

\bibitem[{Sun et~al.(2019)Sun, Gong, Wu, Gong, Jiang, Lan, Sun, and
  Duan}]{sun2019joint}
Sun, C.; Gong, Y.; Wu, Y.; Gong, M.; Jiang, D.; Lan, M.; Sun, S.; and Duan, N.
  2019.
\newblock Joint Type Inference on Entities and Relations via Graph
  Convolutional Networks.
\newblock In \emph{Proceedings of ACL}, 1361--1370.

\bibitem[{Tang et~al.(2020)Tang, Ji, Li, and Zhou}]{tang2020dependency}
Tang, H.; Ji, D.; Li, C.; and Zhou, Q. 2020.
\newblock Dependency Graph Enhanced Dual-transformer Structure for Aspect-based
  Sentiment Classification.
\newblock In \emph{Proceedings of ACL}.

\bibitem[{Vaswani et~al.(2017)Vaswani, Shazeer, Parmar, Uszkoreit, Jones,
  Gomez, Kaiser, and Polosukhin}]{vaswani2017attention}
Vaswani, A.; Shazeer, N.; Parmar, N.; Uszkoreit, J.; Jones, L.; Gomez, A.~N.;
  Kaiser, {\L}.; and Polosukhin, I. 2017.
\newblock Attention is all you need.
\newblock In \emph{NeurIPS}, 5998--6008.

\bibitem[{Walker et~al.(2006)Walker, Strassel, Medero, and
  Maeda}]{walker2006ace}
Walker, C.; Strassel, S.; Medero, J.; and Maeda, K. 2006.
\newblock ACE 2005 multilingual training corpus.
\newblock \emph{Linguistic Data Consortium, Philadelphia} 57.

\bibitem[{Wang et~al.(2018)Wang, Han, Lin, Liu, and
  Sun}]{wang-etal-2018-adversarial}
Wang, X.; Han, X.; Lin, Y.; Liu, Z.; and Sun, M. 2018.
\newblock Adversarial Multi-lingual Neural Relation Extraction.
\newblock In \emph{Proceedings of COLING}, 1156--1166.

\bibitem[{Wang et~al.(2019)Wang, Tu, Wang, and Shi}]{wang-etal-2019-self}
Wang, X.; Tu, Z.; Wang, L.; and Shi, S. 2019.
\newblock Self-Attention with Structural Position Representations.
\newblock In \emph{Proceedings of EMNLP-IJCNLP}, 1403--1409.

\bibitem[{Yang and Mitchell(2016)}]{yang-mitchell-2016-joint}
Yang, B.; and Mitchell, T.~M. 2016.
\newblock Joint Extraction of Events and Entities within a Document Context.
\newblock In \emph{Proceedings of NAACL}, 289--299.

\bibitem[{Zeng et~al.(2014)Zeng, Liu, Lai, Zhou, and
  Zhao}]{zeng-etal-2014-relation}
Zeng, D.; Liu, K.; Lai, S.; Zhou, G.; and Zhao, J. 2014.
\newblock Relation Classification via Convolutional Deep Neural Network.
\newblock In \emph{Proceedings of COLING}, 2335--2344.

\bibitem[{Zhang, Li, and Song(2019)}]{zhang-etal-2019-aspect}
Zhang, C.; Li, Q.; and Song, D. 2019.
\newblock Aspect-based Sentiment Classification with Aspect-specific Graph
  Convolutional Networks.
\newblock In \emph{Proceedings of EMNLP-IJCNLP}.

\bibitem[{Zhang et~al.(2018)Zhang, Zhou, Hong, Yao, and Zhang}]{zhang2018using}
Zhang, J.; Zhou, W.; Hong, Y.; Yao, J.; and Zhang, M. 2018.
\newblock Using entity relation to improve event detection via attention
  mechanism.
\newblock In \emph{CCF International Conference on NLPCC}.

\bibitem[{Zhang, Ji, and Sil(2019)}]{zhang2019joint}
Zhang, T.; Ji, H.; and Sil, A. 2019.
\newblock Joint entity and event extraction with generative adversarial
  imitation learning.
\newblock \emph{Data Intelligence} 99--120.

\bibitem[{Zhang, Qi, and Manning(2018)}]{zhang-etal-2018-graph}
Zhang, Y.; Qi, P.; and Manning, C.~D. 2018.
\newblock Graph Convolution over Pruned Dependency Trees Improves Relation
  Extraction.
\newblock In \emph{Proceedings of EMNLP}, 2205--2215.

\bibitem[{Zhu et~al.(2014)Zhu, Li, Zhou, and Xia}]{zhu-etal-2014-bilingual}
Zhu, Z.; Li, S.; Zhou, G.; and Xia, R. 2014.
\newblock Bilingual Event Extraction: a Case Study on Trigger Type
  Determination.
\newblock In \emph{Proceedings of ACL}, 842--847.

\bibitem[{Zou et~al.(2018)Zou, Xu, Hong, and Zhou}]{zou-etal-2018-adversarial}
Zou, B.; Xu, Z.; Hong, Y.; and Zhou, G. 2018.
\newblock Adversarial Feature Adaptation for Cross-lingual Relation
  Classification.
\newblock In \emph{Proceedings of COLING}, 437--448.

\end{thebibliography}

\appendix






\begin{table*}[!ht]
\centering
\def\arraystretch{1.1}%
\begin{tabular}{l|c|c|c}
\hline
 & English & Chinese & Arabic \\ 
\hline
Relations Mentions & 8,738 & 9,317 & 4,731  \\
Event Mentions & 5,349 & 3,333 & 2,270 \\
Event Arguments & 9,793 & 8,032 & 4,975 \\
\hline
\end{tabular}
\caption{Statistics of the ACE 2005 dataset.
}
\label{table:statistics}
\end{table*}

\begin{table*}[!ht]
\centering
\begin{tabular}{l|c c c|c c c}
\hline
Language & \multicolumn{3}{c}{Sequential Distance} & \multicolumn{3}{c}{Structural Distance} \\ 
\cline{2-7}
& English & Chinese & Arabic & English & Chinese & Arabic \\ 
\hline
Relation mentions & 4.8 & 3.9 & 25.8 & 2.2 & 2.6 & 5.1  \\
Event mentions and arguments & 9.8 & 21.7 & 58.1 & 3.1 & 4.6 & 12.3 \\
\hline
\end{tabular}
\caption{
Average sequential and structural (shortest path) distance between relation mentions and event mentions and their candidate arguments in ACE05 dataset.
Distances are computed by ignoring the order of mentions.
}
\label{table:dist_stat}
\end{table*}
\begin{table*}[!ht]
\centering
\begin{tabular}{l|c|c|c|c}
\hline
Hyper-parameter & \bone & \btwo & \bthree & GATE \\ 
\hline
word embedding size & 300 & 768 & 768 & 768 \\
part-of-speech embedding size & 30 & 30 & 30 & 30 \\
entity type embedding size & 30 & 30 & 30 & 30 \\
dependency relation embedding size & 30 & 30 & 30 & 30 \\
encoder type & GCN & GCN & BiLSTM & Self-Attention \\
encoder layers & 2 & 2 & 1 & 1 \\
encoder hidden size & 200 & 200 & 300 & 512 \\
pooling function & max-pool & max-pool & max-pool & max-pool \\
mlp layers & 2 & 2 & 2 & 2 \\
dropout & 0.5 & 0.5 & 0.5 & 0.5 \\
optimizer & Adam & SGD & Adam & SGD \\
learning rate & 0.001 & 0.1 & 0.001 & 0.1 \\
learning rate decay & 0.9 & 0.9 & 0.9 & 0.9 \\
decay start epoch & 5 & 5 & 5 & 5 \\
batch size & 50 & 50 & 50 & 50 \\
maximum gradient norm & 5.0 & 5.0 & 5.0 & 5.0 \\
\hline
\end{tabular}
\caption{
Hyper-parameters of {\bone} \cite{liu2019neural}, {\btwo} \cite{subburathinam2019cross}, {\bthree} \cite{florian2019neural}, and our approach, GATE.
}
\label{table:hyperparameters}
\end{table*}

\section{Dataset Details}
We conduct experiments on the ACE 2005 dataset which is publicly available from download.\footnote{https://catalog.ldc.upenn.edu/LDC2006T06}
We list the dataset statistics in Table \ref{table:statistics}. 
In Table \ref{table:dist_stat}, we present the statistics of sequential and shortest path distances between relations mentions and event mentions and their arguments in ACE05.

\section{Hyper-parameter Details}

We detail the hyper-parameters for all the baselines and our approach in Table \ref{table:hyperparameters}.

\section{Tuning $\delta$ (shown in Eq. (3))}
\begin{table*}[ht!]
\centering
\begin{tabular}{l|c c c c c c|c}
\hline
\multirow{1}{*}{$\delta$ for Attention Heads} & En $\Rightarrow$ Zh & En $\Rightarrow$ Ar & Zh $\Rightarrow$ En & Zh $\Rightarrow$ Ar & Ar $\Rightarrow$ En & Ar $\Rightarrow$ Zh & Avg. \\ 
\hline
\multicolumn{5}{l}{{Event Argument Role Labeling}} \\
\hline
$[1, 1, 1, 1, \infty, \infty, \infty, \infty]$ & 63.1 & 65.9 & 57.3 & 67.1 & 53.5 & 57.2 & 60.7 \\
$[2, 2, 2, 2, \infty, \infty, \infty, \infty]$ & 64.3 & 69.6 & 58.9 & 69.4 & 52.7 & 56.2 & 61.9 \\
$[4, 4, 4, 4, \infty, \infty, \infty, \infty]$ & 62.1 & 69.8 & 58.9 & 70.5 & 53.0 & 56.1 & 61.7 \\
$[8, 8, 8, 8, \infty, \infty, \infty, \infty]$ & 63.6 & 69.4 & 57.9 & \textbf{71.4} & \textbf{54.0} & 54.9 & 61.9 \\
$[1, 1, 2, 2, \infty, \infty, \infty, \infty]$ & 63.2 & 68.5 & 58.7 & 69.5 & 52.7 & 53.7 & 61.1 \\
$[2, 2, 4, 4, \infty, \infty, \infty, \infty]$ & \textbf{65.0} & 69.6 & \textbf{60.2} & 69.2 & 53.9 & \textbf{57.8} & \textbf{62.6} \\
$[4, 4, 8, 8, \infty, \infty, \infty, \infty]$ & 63.6 & \textbf{70.5} & 58.3 & 70.8 & 53.4 & 57.6 & 62.4 \\
$[1, 2, 4, 8, \infty, \infty, \infty, \infty]$ & 64.3 & 69.6 & 57.8 & 69.7 & 52.5 & 55.5 & 61.6 \\
\hline
\multicolumn{5}{l}{{Relation Extraction}} \\
\hline
$[1, 1, 1, 1, \infty, \infty, \infty, \infty]$ & 54.8 & 63.7 & 70.7	& 62.3 & \textbf{69.8} & 50.6 & 62.0 \\
$[2, 2, 2, 2, \infty, \infty, \infty, \infty]$ & 55.1 & 64.1 & 70.4	& 59.4 & 68.7 & 50.2 & 61.3 \\
$[4, 4, 4, 4, \infty, \infty, \infty, \infty]$ & 55.5 & 64.5 & \textbf{71.6}	& 61.2 & 68.7 & 51.5 & 62.2 \\
$[8, 8, 8, 8, \infty, \infty, \infty, \infty]$ & 55.5 &	\textbf{65.5} & 71.1 & 61.7 & 67.5 & \textbf{53.4} & \textbf{62.5} \\
$[1, 1, 2, 2, \infty, \infty, \infty, \infty]$ & 56.4 & 63.5 & 70.4	& \textbf{63.1} & 69.4 &	51.9 & \textbf{62.5} \\
$[2, 2, 4, 4, \infty, \infty, \infty, \infty]$ & 55.6 & 62.0 & 70.6	& 61.6 & 67.2 & 51.2 & 61.4 \\
$[4, 4, 8, 8, \infty, \infty, \infty, \infty]$ & \textbf{55.8} & 63.9 & 71.5	& 63.0 & 68.5 & 50.6 & 62.2 \\
$[1, 2, 4, 8, \infty, \infty, \infty, \infty]$ & 55.4 & 65.0 & 70.3	& 61.1 & 69.6 &	50.7 & 62.0 \\
\hline
\end{tabular}
\caption{
Event Argument Role Labeling (EARL) and Relation Extraction (RE) \emph{single-source transfer} results (F-score \%) of our proposed approach GATE with different distance threshold $\delta$ using perfect event triggers and entity mentions.
En, Zh, and Ar denotes English, Chinese, and Arabic languages, respectively.
In ``X $\Rightarrow$ Y'', X and Y denotes the source and target language, respectively.
}
\label{table:delta_tuning}
\end{table*}

During our initial experiments, we observed that setting $\delta = \infty$ in four attention heads provide consistently better performances. 
We tune $\delta$ in the range $[1, 2, 4, 8]$ on the validation set based on the statistics of the shortest path distances between relations mentions and event mentions and their arguments in ACE05 (shown in Table \ref{table:dist_stat}).
we set $\delta = [2, 2, 4, 4, \infty, \infty, \infty, \infty]$ and $\delta = [1, 1, 2, 2, \infty, \infty, \infty, \infty]$ for the event argument role labeling and relation extraction tasks, respectively, in all our experiments.
This hyper-parameter choice provides us comparably better performances (on test sets), as shown in Table \ref{table:delta_tuning}.



\section{Translation Experiment}
\begin{figure*}[t]
\centering
\includegraphics[width=0.8\linewidth]{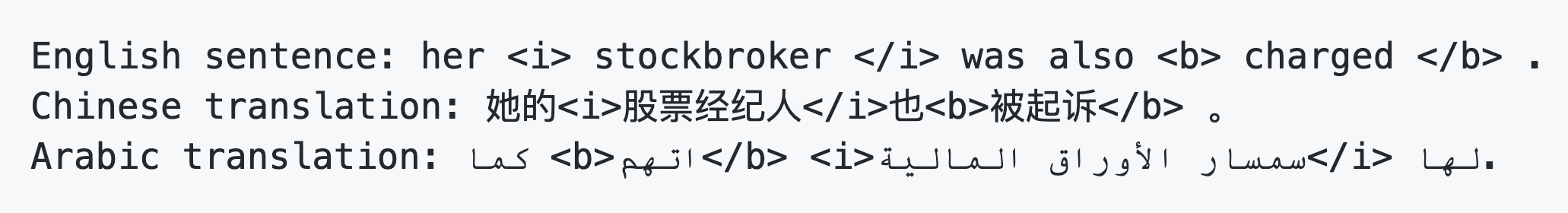}
\caption{
Translation of an English sentence in Chinese and Arabic with an event trigger (surrounded by {\textless b\textgreater}{\textless /b\textgreater}) and a candidate argument (surrounded by {\textless i\textgreater}{\textless /i\textgreater}).
}
\label{fig:trans_ex}
\end{figure*}

We perform English to Arabic and Chinese translations using {\tt Google Cloud Translate}.\footnote{https://cloud.google.com/translate/docs/basic/setup-basic}
During translation, we use special symbols to identify relation mentions and event mentions and their argument candidates in the sentences, as shown in Figure \ref{fig:trans_ex}.
We drop the examples ($\approx 10\%$) in which we cannot identify the mentions after translation.

\begin{table*}[!ht]
\centering
\begin{tabular}{l|c c c c}
\hline
\multirow{1}{*}{Model} & True Positive & True Negative & False Positive & False Negative \\
\hline
Self-Attention  & 386 & 563 & 179 & 300 \\
GATE  & 585 & 493 & 249 & 157 \\
\hline
\end{tabular}
\caption{
Comparing GATE and Self-Attention on the EARL task using English and Chinese as the source and target languages, respectively.
The rates are aggregated from confusion matrices shown in Figure \ref{fig:gate_event_cm} and  \ref{fig:tran_event_cm}.
}
\label{table:pred_summary}
\end{table*}

\section{Error Analysis}

\begin{figure*}[t]
\centering
\includegraphics[width=1.0\linewidth]{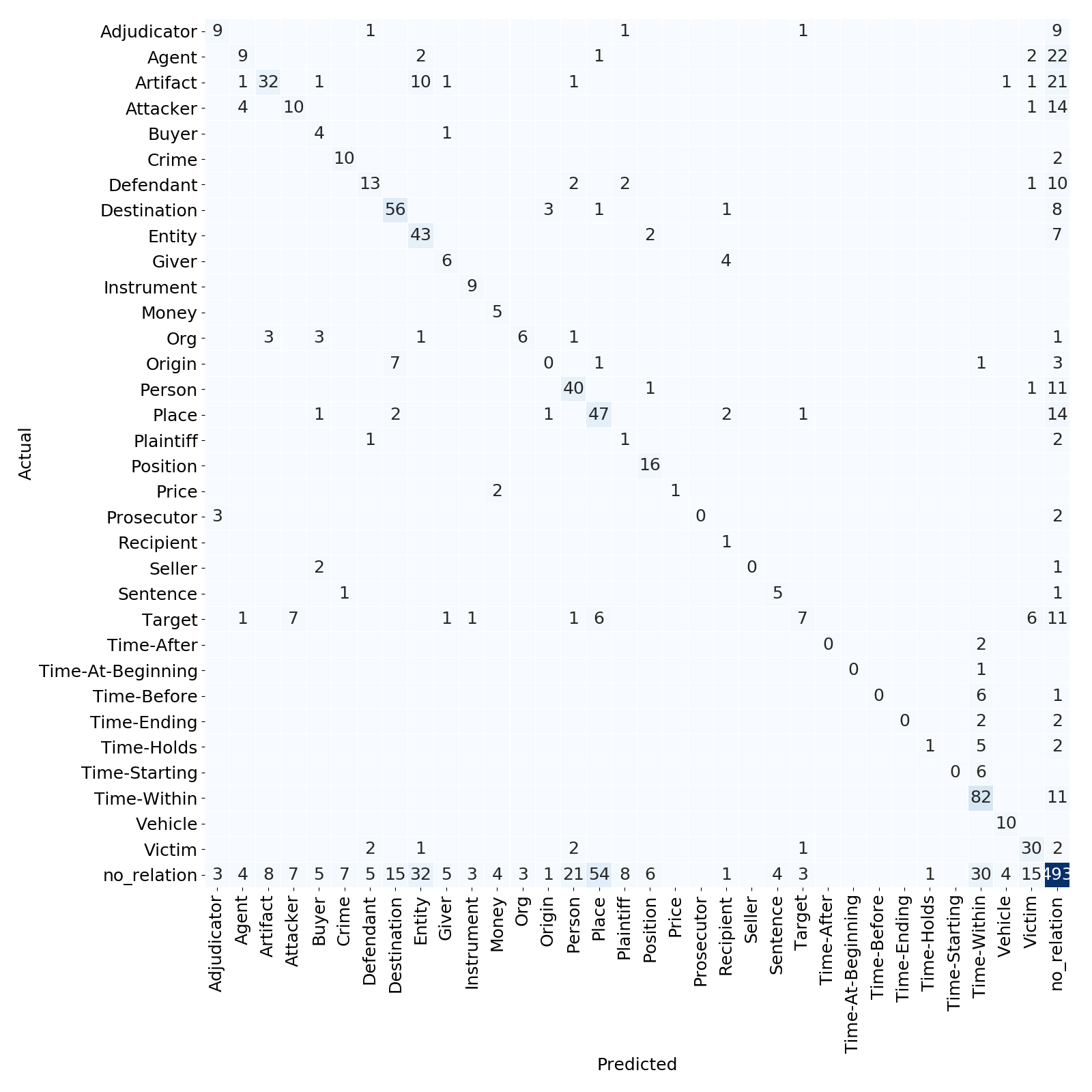}
\caption{
Event argument role labeling confusion matrix (on test set) based on our proposed approach \textbf{GATE} using English and Chinese as the source and target languages, respectively.
The diagonal values indicate the number of correct predictions, while the other values denote the incorrect prediction counts.
}
\label{fig:gate_event_cm}
\end{figure*}

\begin{figure*}[t]
\centering
\includegraphics[width=1.0\linewidth]{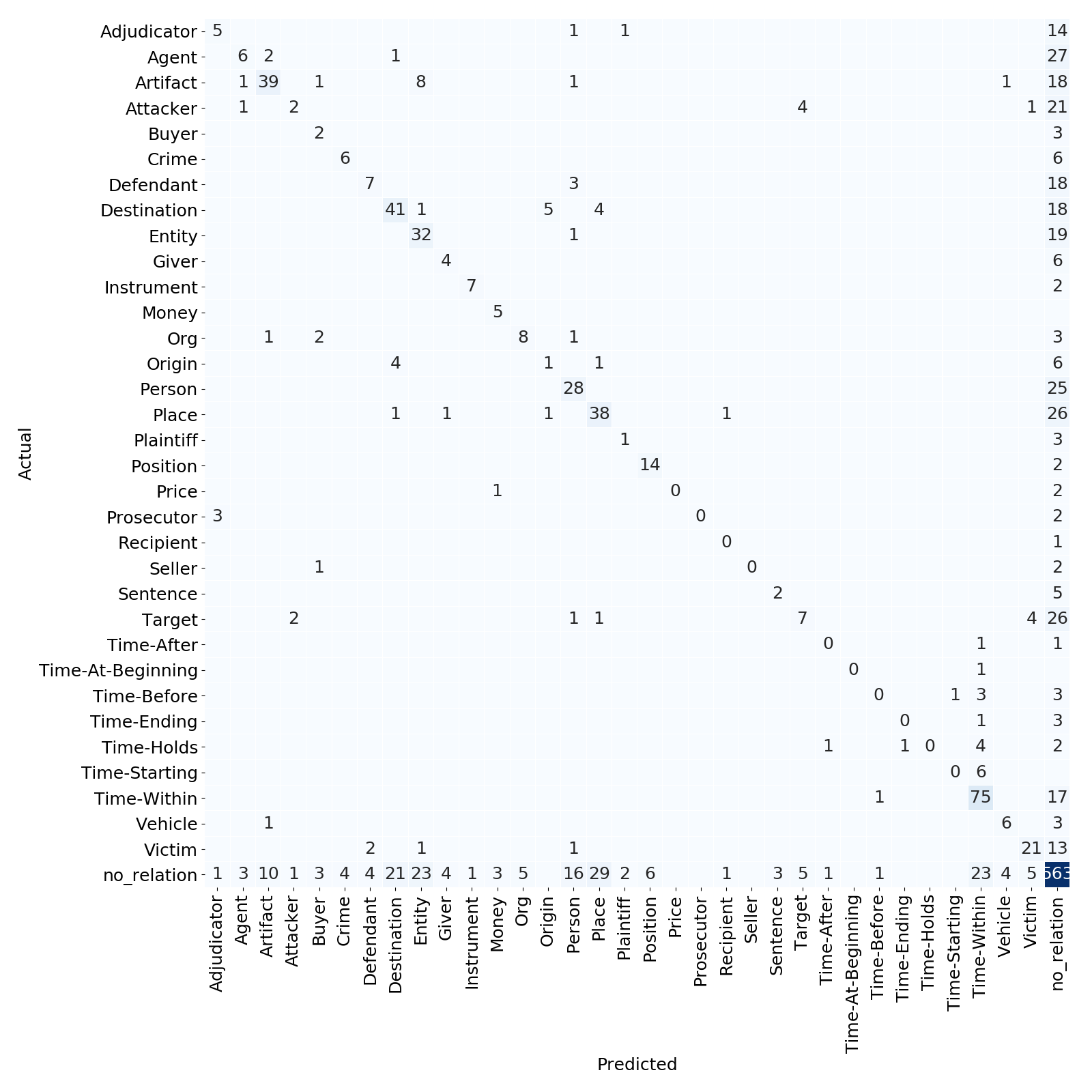}
\caption{
Event argument role labeling confusion matrix (on test set) based on the \textbf{Self-Attention (Transformer Encoder)} using English and Chinese as the source and target languages, respectively.
The diagonal values indicate the number of correct predictions, while the other values denote the incorrect prediction counts.
}
\label{fig:tran_event_cm}
\end{figure*}

\begin{figure*}[t]
\centering
\includegraphics[width=1.0\linewidth]{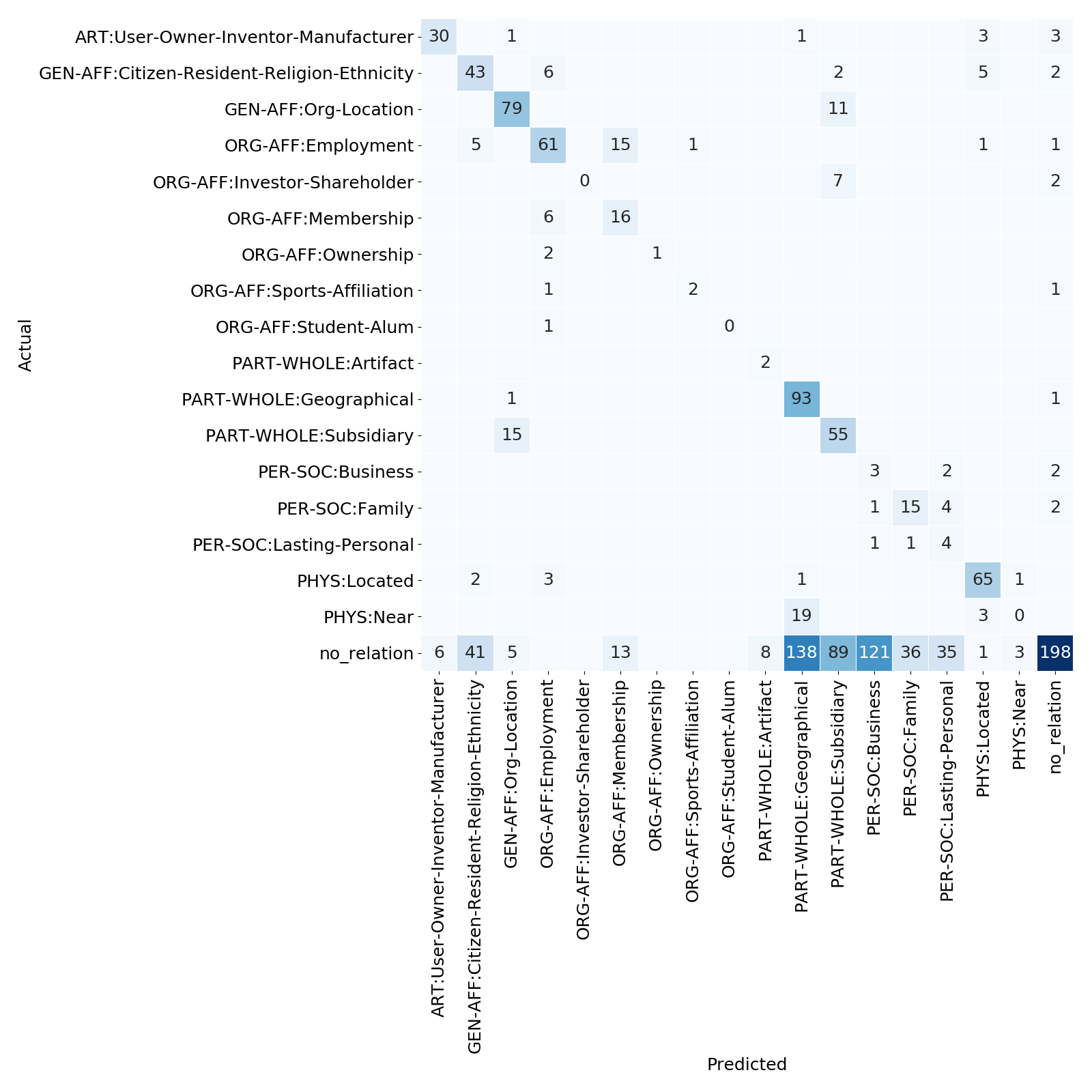}
\caption{
Relation extraction labeling confusion matrix (on test set) based on our proposed approach \textbf{GATE} using English and Chinese as the source and target languages, respectively.
The diagonal values indicate the number of correct predictions, while the other values denote the incorrect prediction counts.
}
\label{fig:gate_relation_cm}
\end{figure*}

\begin{figure*}[t]
\centering
\includegraphics[width=1.0\linewidth]{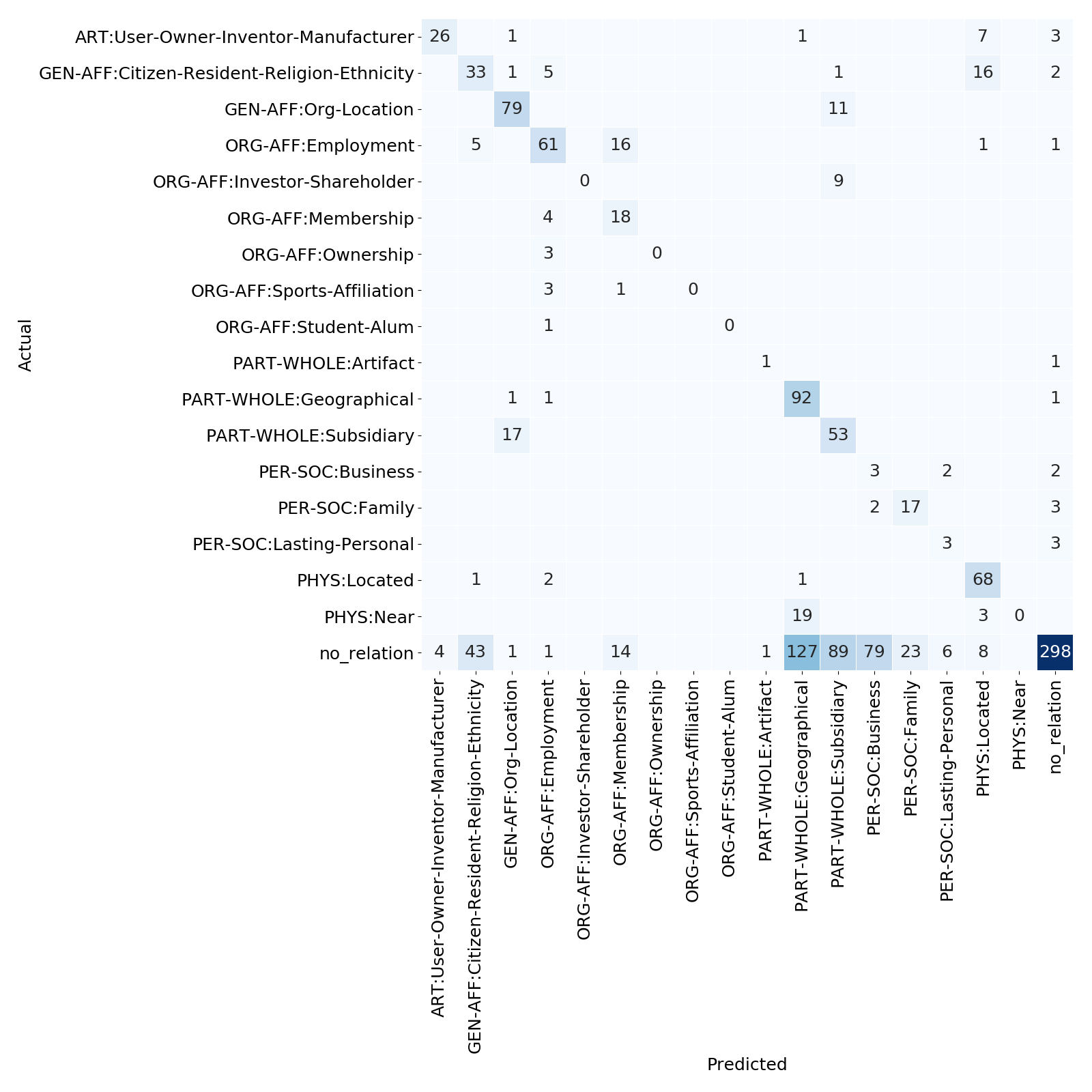}
\caption{
Relation extraction confusion matrix (on test set) based on the \textbf{Self-Attention (Transformer Encoder)} using English and Chinese as the source and target languages, respectively.
The diagonal values indicate the number of correct predictions, while the other values denote the incorrect prediction counts.
}
\label{fig:tran_relation_cm}
\end{figure*}

We compare our proposed approach GATE and the self-attention mechanism \cite{vaswani2017attention} on the event argument role labeling (EARL) and relation extraction (RE) tasks. 
We consider the models trained on English language and evaluate them on Chinese language.
We do not use the event trigger type as features while training models for the EARL task.
We present the confusion matrices of these two models in Figure \ref{fig:gate_event_cm}, \ref{fig:tran_event_cm}, \ref{fig:gate_relation_cm}, and \ref{fig:tran_relation_cm}.
In general, GATE makes more correct predictions. 
We noticed that in transferring from English to Chinese on the EARL task, GATE improves notably on \textsf{Destination, Entity, Person, Place} relation types.
The syntactic distance between event triggers and their argument mentions that share those types corroborates with our hypothesis that distance-based dependency relations help in cross-lingual transfer.

However, we observed that GATE makes more \emph{false positive} and less \emph{false negative} predictions than the self-attention mechanism. 
We summarize the prediction rates on EARL in Table \ref{table:pred_summary}.
There are several factors that may be associated with these wrong predictions.
To shed light on those factors, we manually inspect 50 examples and our findings suggests that wrong predictions are due to three primary reasons. 
First, there are errors in the ground truth annotations in the ACE dataset. 
Second, the knowledge required for prediction is not available in the input sentence.
Third, there are entity mentions, event triggers, and contextual phrases in the test data that rarely appear in the training data.

\end{document}